\documentclass{article}
\usepackage{ijcai23}

\usepackage{times}
\usepackage{soul}
\usepackage{url}
\usepackage[hidelinks]{hyperref}
\usepackage[utf8]{inputenc}
\usepackage[small]{caption}
\usepackage{graphicx}
\usepackage{amsmath}
\usepackage{amsthm}
\usepackage{booktabs}
\usepackage{algorithmic}
\usepackage[switch]{lineno}

\usepackage[ruled, linesnumbered]{algorithm2e}
\usepackage{algorithmic}

\usepackage{amsmath}
\usepackage{bm}
\usepackage{mathrsfs}
\usepackage{amsthm}
\usepackage{amsfonts}
\usepackage{subfigure}
\usepackage{multirow}
\usepackage{url}
\usepackage{array}
\usepackage{amssymb}
\usepackage{booktabs} 

\usepackage{natbib}

\usepackage{color} 

\title{\textsc{Congregate}: Contrastive Graph Clustering in Curvature Spaces}

\author{
    Li Sun$^1$ \and
    Feiyang Wang$^2$ \and
    Junda Ye$^2$ \and
    Hao Peng$^3$ \and
    Philip S. Yu$^4$
\affiliations
    $^1$North China Electric Power University, Beijing 102206, China\\
    $^2$Beijing University of Posts and Telecommunications, Beijing 100876, China\\
    $^3$Beihang University, Beijing 100191, China\\
    $^4$Department of Computer Science, University of Illinois at Chicago, IL, USA
\emails
    ccesunli@ncepu.edu.cn; \{fywang, jundaye\}@bupt.edu.cn; penghao@buaa.edu.cn; psyu@uic.edu
}

\begin{document}

\maketitle

\begin{abstract}
Graph clustering is a longstanding research topic,
and has achieved remarkable success with the deep learning methods in recent years. 
Nevertheless, we observe that several important issues largely remain open.
On the one hand, graph clustering from the geometric perspective is appealing but has rarely been touched before, as it lacks a promising space for geometric clustering.
On the other hand, contrastive learning boosts the deep graph clustering but usually struggles in either graph augmentation or hard sample mining.
To bridge this gap, we rethink the problem of graph clustering from geometric perspective and,
to the best of our knowledge, make the first attempt to \emph{introduce a heterogeneous curvature space to graph clustering problem}.
Correspondingly, we present a novel end-to-end contrastive graph clustering model named \textsc{Congregate}, addressing  geometric graph clustering with Ricci curvatures.
To support geometric clustering, we construct a theoretically grounded Heterogeneous Curvature Space where deep representations are generated via the product of the proposed \emph{fully Riemannian} graph convolutional nets.
Thereafter, 
we train the graph clusters by an \emph{augmentation-free} reweighted contrastive approach where we pay more attention to \emph{both hard negatives and hard positives} in our curvature space.
Empirical results on real-world graphs show that our model outperforms the state-of-the-art competitors.
\end{abstract}


\section{Introduction}

Graph clustering aims to group nodes into different clusters so that 
the intra-cluster nodes share higher similarity than the inter-cluster ones,
receiving continuous research attention \citep{DBLP:conf/kdd/YinBLG17}.
The state-of-the-art clustering performance on graphs has been achieved by deep clustering methods in recent years \citep{DBLP:journals/corr/abs-2212-08665,DBLP:conf/ijcai/WangPHLJZ19,DBLP:conf/nips/LiYLZZR0H20}.
Meanwhile, we find that several important issues on deep graph clustering still largely remain open.

\emph{The first issue is on the geometric graph clustering.}
In the literature, classic concepts such as modularity \citep{DBLP:conf/www/LiJT22}, conductance \citep{DBLP:conf/cikm/DuvalM22} and motifs \citep{DBLP:conf/www/JiaZ0W19} are frequently revisited.
Little effort has been devoted to clustering from a geometric perspective.
In the Riemannian geometry, \textbf{Ricci curvatures} on the edges 
can help determine the cluster boundary \citep{DBLP:journals/dcg/JostL14}, 
thereby showing the density and clustering behavior among the nodes.
However, graph clustering has rarely been touched yet in Riemannian geometry,
since \emph{it lacks a promising Riemannian space for graph clustering}.
Most existing graph representation spaces present a single curvature radius, independent of nodes/edges \citep{pRieGNN,HGCN,NEURIPS2021_b91b1fac}, and cannot allow for a closer look over the various curvatures for graph clustering.
Also, typical clustering algorithms in Euclidean space  (e.g., $K$-means) cannot be directly applied as an alternative, due to the inherent difference in geometry.
Consequently, it calls for a new Riemannian curvature space, \emph{supporting a fine-grained curvature modeling for geometric clustering}.

The second is on the unsupervised learning.
Deep models are typically trained by the supervisions while
graph clustering is unsupervised by nature.
Recently, the contrastive clustering without external supervision draws dramatic attention \citep{DBLP:journals/corr/abs-2204-08504,S3GC,DBLP:conf/www/LiJT22}.
\emph{In the line of contrastive graph clustering, the issues of augmentation and hard samples are still unclear in general.}
Unlike the easily obtained augmentations on images, graph augmentation is nontrivial \citep{HassaniA20}.
In addition, the noise injected in this process usually requires a careful treatment to avoid misleading on graph clustering \citep{DBLP:conf/ijcai/GongZTL22}.
\cite{DBLP:conf/iclr/RobinsonCSJ21} point out the hardness unawareness of typical loss function such as InfoNCE.
Hard negative samples have shown to be effective for graph contrastive learning \citep{DBLP:conf/icml/XiaWWCL22},
but little effort is made to its counterpart, \emph{hard positive samples}.
In fact, the hard positives in our context are the border nodes of a cluster, and plays a crucial role in clustering performance.
Unfortunately, hard sample mining in curvature space largely remains open.

Motivated by the observations above, we rethink the problem of graph clustering from the geometric perspective, and make the first attempt to address graph clustering in a  novel \textbf{Curvature Space}, \emph{rather than traditional single curvature ones}, with an advanced contrastive loss.

\textbf{Our Work.} To this end, we propose 
a novel end-to-end \underline{con}trastive \underline{gr}aph clust\underline{e}rin\underline{g} model in curv\underline{at}ure spac\underline{e}s (\textsc{Congregate}),
where we approach graph clustering via geometric clustering with Ricci curvatures
so that positive Ricci curvature groups the nodes while negative Ricci departs them in spirit of the famous Ricci flow. 
To address the fine-grained curvature modeling for graph clustering (\emph{the first issue}),
we introduce a novel \emph{Heterogeneous Curvature Space},  which is a key innovation of our work.
It is designed as the product of learnable factor manifolds and multiple free coordinates.
We prove that the proposed space allows for different curvatures on different regions, 
and the fine-grained node curvatures can be inferred to accomplish curvature modeling.
Accordingly, we generate deep representations  via the product of Graph Convolutional Nets (GCNs), 
where  \emph{fully Riemannian} GCN is designed to address the inferior caused by tangent spaces.
To address the unsupervised learning (\emph{the second issue}), we propose a rewighted geometric contrastive approach in our curvature space.
On the one hand,  our approach is \emph{free of augmentation} as 
we contrast across the geometric views generated from the proposed heterogeneous curvature space  itself.
On the other hand, we equip a novel dual reweighting to the Node-to-Node and Node-to-Cluster contrastive losses to train the clusters.
In this way, we pay more attention to \emph{both hard negatives and hard positives} when
maximizing intra-cluster similarity and minimizing inter-cluster similarity.


To sum up, the noteworthy contributions are listed below:
\vspace{-0.03in}
\begin{itemize}
   \item \emph{Problem}.  We rethink the graph clustering from geometric respective. To the best of our knowledge, we are the first to introduce the heterogeneous curvature space, supporting fine-grained curvatures modeling, to the problem of graph clustering.
  \vspace{-0.03in}
    \item \emph{Methodology}. We propose an end-to-end \textsc{Congregate} free of graph augmentation,  in which we approach geometric graph clustering with the reweighting contrastive loss in the proposed \emph{heterogeneous curvature space}, paying attention to hard positives and hard negatives.
    \vspace{-0.01in}
\item \emph{Experiments}. We evaluate the superiority of our model with $19$ strong competitors, examine the proposed components by ablation study, and further discuss why Ricci curvature works.
\end{itemize}

\vspace{-0.18in}
\section{Preliminaries}
\vspace{-0.03in}
In this section, we first introduce the necessary fundamentals of Riemannian geometry for better understanding our work, and then formulate the studied problem in this paper.
In short, \emph{we are interested in the end-to-end graph clustering in a novel curvature space}.

\vspace{-0.07in}
\subsection{Riemannian Geometry}
\vspace{-0.01in}

\noindent \textbf{Manifold.}
A Riemannian manifold $(\mathcal M, g)$ is a smooth manifold $\mathcal M$ endowed with a Riemannian metric $g$. 
Every point $x \in \mathcal M$ is associated with a Euclidean-like \emph{tangent space} $\mathcal T_x\mathcal M$ on which the metric $g$ is defined.
The \emph{exponential map} projects from the tangent space onto the manifold, and the \emph{logarithmic map} does inversely \citep{2013manifold}.


\vspace{0.02in}
\noindent \textbf{Curvature.}
For each point $x$ in the manifold, it is coupled with a curvature $c_x$ describing how the space around $x$ derives from being flat and a corresponding curvature radius $\frac{1}{|c_x|}$.
When $c_x$ is equal everywhere in the manifold, 
it induces a \textbf{homogeneous} curvature space (a.k.a. constant curvature space) with a simplified notation of scalar curvature $c$. 
Concretely, 
it is said to be \emph{hyperbolic}  $\mathbb H$ if $c<0$, and \emph{hyperspherical} $\mathbb S$  if $c>0$. \emph{Euclidean} space $\mathbb R$ is special case with $c=0$.
On the contrary, \textbf{heterogeneous} curvature space refers to a manifold whose curvatures on different regions are not the same, which is a more practical yet challenging case.

\vspace{-0.09in}
\subsection{Problem Formulation}
\vspace{-0.01in}
In this paper, we consider the node clustering on attributed graphs. 
An attributed graph is described as a triplet of $G=(\mathcal V, \mathcal E, \mathbf X)$, 
where $\mathcal V=\{v_1, v_2, \cdots, v_N \}$  is the set of $N$ nodes, $\mathcal E \subset \mathcal V \times \mathcal V$  is the edge set, and $\mathbf X \in \mathbb R^{N \times F}$ is the attribute matrix.
Let $K$ denote the number of node clusters. 
The node-to-cluster assignment is described as the \emph{cluster membership} vector $\boldsymbol \pi_i \in \mathbb R^K$ attached to node $v_i$. 
$\boldsymbol \pi_i$ is a stochastic vector adding up to $1$, whose $k$-th element $\pi_{ik}$ is the probability of $v_i$ belonging to cluster $k$.
Now, we formulate the problem of \underline{Geometric Graph Clustering in Generic Curvature Space}. 
\vspace{-0.012in}
\newtheorem*{def0}{Problem Definition} 
\begin{def0}
Given $G=(\mathcal V, \mathcal E, \mathbf X)$, the goal of our problem is to learn an encoder $f: v_i \to [\boldsymbol z_i, \boldsymbol \pi_i], \forall v \in \mathcal V$ that 1) directly outputs cluster membership $\boldsymbol \pi_i$  \emph{(end-to-end)} so that the nodes are more similar to those grouped in the same cluster than the nodes in different clusters and 2) the node encodings in the generic curvature space $\boldsymbol z_i \in \mathcal M$, \emph{supporting the geometric graph clustering}.
\end{def0}
\vspace{-0.03in}
\noindent Distinguishing with the prior works, we rethink the problem of graph clustering from the geometric perspective, and make the first attempt to study graph clustering in a novel \emph{Curvature Space}, rather than traditional single curvature ones.

\vspace{0.03in}
\noindent \textbf{Notations.} The lowercase $x$,  boldfaced $\boldsymbol x$ and uppercase $\mathbf X$ denote scalar, vector and matrix, respectively.

\begin{figure*}
\centering
     \vspace{-0.16in}
    \includegraphics[width=1\linewidth]{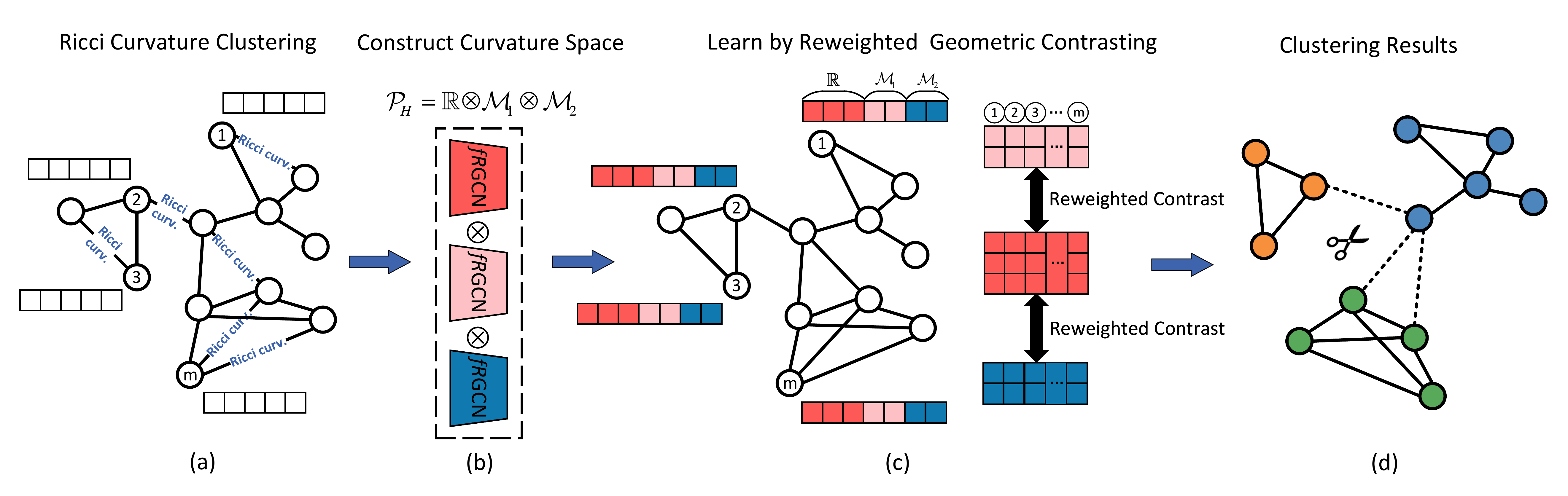}
           \vspace{-0.32in}
     \caption{Illustration of \textsc{Congregate}. (a) We address graph clustering from geometric perspective with Ricci curvatures. (b) We construct a novel curvature space where we generate deep representations via the product of  proposed \emph{\textbf{f}RGCN}s. (c) Our model is trained by a reweighted contrastive loss across geometric views (\textcolor{red}{red}/\textcolor{magenta}{magenta}/\textcolor{blue}{blue}) free of augmentation. (d) We obtain clustering results in an end-to-end fashion. }
    \label{illu}
    \vspace{-0.18in}
\end{figure*}

\vspace{-0.12in}
\section{Methodology: \textsc{Congregate}}
\vspace{-0.03in}
We propose
an end-to-end contrastive graph clustering model (\textsc{Congregate}) where \emph{we introduce the first curvature space to graph clustering}, a key innovation of our work.
In brief, we directly learn the node clusters by training randomly initialized centroids $\{\boldsymbol \phi_k\}_{k=1,\cdots, K}$  in a novel curvature space.
$\boldsymbol \phi_k$ is the centroid of cluster $k$.
The soft assignment of node $v_i$ to cluster $k$ is  given as $\pi_{ik}=Normalize(\delta(\boldsymbol z_i, \boldsymbol \phi_k))$, where the similarity $\delta(\boldsymbol z_i, \boldsymbol \phi_k)=exp(-d_{\mathcal P}(\boldsymbol z_i, \boldsymbol \phi_k))$  and $d_\mathcal P$ is distance metric in our curvature space.
Softmax normalization is applied so that $\boldsymbol \pi_i$ adds up to $1$.

We illustrate our model in Figure 1. Concretely, we present a  geometric clustering approach with Ricci curvatures (\textbf{Sec \ref{1}}), introduce the novel heterogeneous curvature space (\textbf{Sec \ref{2}}), and train cluster centroids by the proposed reweighted contrastive loss in our curvature space (\textbf{Sec \ref{3}}).

\vspace{-0.08in}
\subsection{Geometric Clustering with Ricci Curvature} \label{1}
\vspace{-0.03in}
In \textsc{Congregate}, we address graph clustering from a geometric perspective, more concretely, \emph{the notion of Ricci curvature}, and formulate a novel geometric clustering loss.

We first discuss why Ricci curvature clusters nodes. Let us begin with its definition \citep{DBLP:journals/dcg/JostL14,2011Ricci}: Given a graph with mass distribution $m^\lambda_i( \cdot)$ on $v_i$’s neighbor nodes, Ricci curvature $Ric(i,j)$ of edge $(v_i, v_j)$  is defined as 
\vspace{-0.07in}
\begin{equation}
Ric(i,j)=1- \frac{W(m^\lambda_i, m^\lambda_j)}{d_G(v_i,v_j)},
\vspace{-0.05in}
\end{equation}
 and $W(m^\lambda_i, m^\lambda_j) $ is the Wasserstein distance between the mass distributions on nodes, where $m^\lambda_i( \cdot)$ is defined as
 \vspace{-0.07in}
\begin{equation}
m_i^{\lambda}\left(v_j \right)= \begin{cases}\lambda & \text { if } v_j=v_i \\  \frac{1-\lambda}{degree_i} & \text { if } v_j \in \mathcal N_i, \end{cases}
\end{equation}
where $d_G$ is the length of shortest path on the graph, and $\lambda$ is a control parameter.
The intuition is that \emph{the Ricci curvature of an edge describes the overlap extent between neighborhoods of its two end nodes, and thus signifies the density among nodes.} 
Specifically, if $v_i$ and $v_j$ belong to different clusters, it is costing to move the distribution $m^\lambda_i$ to $m^\lambda_j$ due to fewer common neighbors. The less overlapped neighborhoods present large $W(m^\lambda_i, m^\lambda_j)$ and negative $Ric(i,j)$. On the contrary, intra-cluster edges are most positively curved, and the nodes within the cluster are densely connected.

With the observation above, we connect the Ricci curvature on edges to the density among the nodes. Then, intra-cluster density is formulated as summing the $Ric(i,j)$ whose end nodes belong to the same cluster,
\vspace{-0.03in}
\begin{equation}
D_{intra}=\frac{1}{|\mathcal E|}\sum\nolimits_{i,j} \sum\nolimits_{k=1}^K Ric(i,j)\pi_{ik}\pi_{jk}.
\vspace{-0.03in}
\label{intra}
\end{equation}
Similarly, the inter-cluster density is given as
\begin{equation}
D_{inter}=\frac{1}{|\mathcal E|K}\sum\nolimits_{i,j} \sum\nolimits_{k_1 \neq k_2}Ric(i,j)\pi_{ik_1}\pi_{jk_2}.
\vspace{-0.05in}
\label{inter}
\end{equation}
Consequently, the Ricci loss is defined as follows,
\begin{equation}
\mathcal L_{Ric}=\alpha_0 D_{inter} -D_{intra},
\label{ricciLoss}
\end{equation}
where $\alpha_0$ is a weighting coefficient. The rationale of our formulation is that we  maximize  node density within the cluster while minimizing the density across different clusters.

\noindent \textbf{Connection to the Famous Ricci Flow.}
In differential geometry, the Ricci flow approach is to divide a smooth manifold into different regions based on the Ricci curvature.
The regions of large positive curvature shrink in whereas regions of very negative curvature spread out \citep{2005Ricci}.
Analogy to the smooth manifold, we divide a graph into different node clusters where \emph{positive Ricci curvature groups the nodes and negative Ricci departs them.}



\citet{CDRiccFlow,ORCurvCD} leverage Ricci curvatures to group nodes, but they do not consider the end-to-end clustering in a curvature space, essentially different from our setting.
We are the first to introduce the curvature space to the problem of graph clustering to the best of out knowledge.

\vspace{-0.07in}
\subsection{Constructing Heterogeneous Curvature Space} \label{2}
\vspace{-0.02in}
We are facing a challenging task: constructing a new curvature space for the geometric graph clustering.
Most existing graph curvature spaces  \emph{present as a single curvature radius} (either the typical hyperbolic, spherical and Euclidean spaces or the recent ultrahyperbolic and quotient manifolds \citep{pRieGNN,NEURIPS2021_b91b1fac}). 
However, rather than a single curvature, geometric clustering requires a closer look over the various fine-grained curvatures on the graph.

A core contribution of our work is that we introduce a novel \emph{heterogeneous curvature space}, bridging this gap.
In a nutshell, it is a product space of learnable factor manifolds and multiple free coordinates, as shown in Fig 1 (b).

\vspace{-0.05in}
\subsubsection{A Novel Product Manifold} \label{3-2}
We introduce the intuition of our idea before the formal theory.
The graph curvature spaces above are restricted by a fixed norm, thus yielding a single curvature radius.
We enrich the curvatures by producting a single radius space with \textbf{multiple free coordinates} that do not have any norm restriction. (A more theoretical rationale based on rotational symmetry \citep{DBLP:journals/corr/abs-2202-01185} is given in Appendix.)
Our heterogeneous curvature space $\mathcal P_H$ is constructed as follows, 
\vspace{-0.05in}
\begin{equation}
\mathcal P_H= \otimes^M_{m=0} \mathcal M_m^{c_m,d_m}, \ \mathcal M_0^{c_0,d_0}:=\mathbb R^{d_0}, c_0=0,
\end{equation}
where $\otimes$ denotes the Cartesian product. 
It is a product of $M$ \emph{restricted factors} and \emph{a free factor of $d_0$ free coordinates}. 
In the product space, 
a point  $\boldsymbol z \in \mathcal P_H$ is thus expressed as the concatenation of its factors
$\boldsymbol z^m \in \mathcal M_m^{c_m,d_m}$ 
with the combinational distance metric of 
$d_{\mathcal P}^2(\boldsymbol x, \boldsymbol y)= \sum\nolimits_m d_{c_m}^2(\boldsymbol x^m, \boldsymbol y^m)$.

\emph{A restricted factor} $\mathcal M_m^{c_m,d_m}$ is defined on the manifold, 
\vspace{-0.05in}
\begin{equation}
\left\{ \boldsymbol z=
\left[\begin{array}{c}
z_t  \\
\boldsymbol z_s
\end{array}\right]
 \bigg|  \ \langle \boldsymbol z, \boldsymbol z \rangle_{c_m} = \frac{1}{c_m}, \ z_t \in \mathbb R, \boldsymbol z_s  \in \mathbb R^{d_m}\right\}, 
\end{equation}
with the metric inner product 
$\langle \boldsymbol z, \boldsymbol z \rangle_{c_m}=sgn(c_m)z^2_t + \boldsymbol z_s^\top\boldsymbol z_s$,
where $sgn$ is the sign function. 
$c_m$ and $d_m$ denote the curvature and dimension, respectively.
The induced \emph{norm restriction} is given as $\| \boldsymbol z \|^2_{c_m} =\langle \boldsymbol z, \boldsymbol z \rangle_{c_m}$.
$z_t$ is the $1$st dimension, and is usually termed as $t$-dimension. 
The north pole is $\mathbf 0 = (|c_m|^{-\frac{1}{2}}, 0, \cdots, 0)$. 
The closed-form distance $d_{c_m}$,
logarithmic $log^{c_m}_{\boldsymbol z}$  and exponential maps $exp^{c_m}_{\boldsymbol z}$ are derived in \cite{SkopekGB20}.
\emph{The free factor}  $\mathbb R^{d_0}$  looks Euclidean like, but in fact we inject the rotational symmetry in it.
The closed-form distance $d_{0}$ is given in \citep{DBLP:journals/corr/abs-2202-01185}.
We do not use its logarithmic/exponential maps in our model.

We prove that the proposed $\mathcal P_H$ has heterogeneous curvatures, i.e., it allows for different curvatures on the different regions. Supporting curvature heterogeneity is the foundation of geometric clustering.
We start with the concept below.

\vspace{-0.02in}
\newtheorem*{def1}{Definition (Diffeomorphism \citep{2013manifold})} 
\begin{def1}
Given two manifolds $\mathcal M_1$ and $\mathcal M_2$, a smooth map $\varphi: \mathcal M_1 \to \mathcal M_2$ is referred to as a diffeomorphism if $\varphi$ is bijective and its inverse $\varphi^{-1}$ is also smooth. $\mathcal M_1$ and $\mathcal M_2$ are said to be diffeomorphic and denoted as $\mathcal M_1 \simeq \mathcal M_2$ if there  exists a $\varphi$ connecting them.
\end{def1}

\vspace{-0.02in}
\newtheorem*{prop1}{Proposition 1 (Curvature Heterogeneity)} 
\begin{prop1}
$\forall d_0 >1, \forall c_m$, there exists a \emph{diffeomorphism} of $\mathcal P_H \simeq (\otimes_{m=1}^M \mathcal M_m^{c_m, d_m} \otimes \mathcal M^{0, d}) \otimes \mathbb R_S$ where a point $\boldsymbol z_i$'s curvature is a map $\psi((\boldsymbol z_i)_{[S]}, $ $c_1, \cdots, c_M)$ w.r.t. its location with the differential operator
  \vspace{-0.03in}
\begin{equation}
\frac{-2\partial^2_{SS}\rho}{\rho} + \frac{1-(\partial^2_{S}\rho)^2}{\rho^2}, 
\end{equation}
for some smooth $\rho$ and $(\boldsymbol z_i)_{[S]}$ is the coordinate of $\mathbb R_S$, where $ \mathcal M^{0, d} \otimes \mathbb R_S = \mathbb R^{d_0}$ and $\mathbb R_S$ is the axis for rotational symmetry.
\end{prop1}
  \vspace{-0.05in}
\begin{proof}
Please refer to the Appendix.
\end{proof}
\vspace{-0.1in}
\subsubsection{Fine-grained Curvature Modeling for Graph Clustering}

Here, we derive the fine-grained node-level curvature in our product space.
With the definition of \emph{Diffeomorphism} above and \emph{Proposition 1},  the curvature $c_i$ of  $\boldsymbol z_i \in \mathcal P_H$ can be derived from the map  $(\varphi \circ \psi)((\boldsymbol z_i)_{[S]}, c_1, \cdots, c_M)$ and the differential operator on $\rho$. 
That is, $\boldsymbol z_i$'s curvature is inferred via a map regarding the curvatures of factor manifold $c_1, \cdots, c_M$ and its coordinate of rotation symmetry $(\boldsymbol z_i)_{[S]}$.
In our construction, $(\boldsymbol z_i)_{[S]}$ is given in the $1$st dimension of the $\boldsymbol z_i$'s free factor $(\boldsymbol z_i^{0})_{[1]}$.
We employ a multilayer perceptron (MLP) to approximate the map.
The estimated curvature $\bar{c}_i$ is given as,
  \vspace{-0.1in}
\begin{equation}
\bar{c}_i=MLP([(\boldsymbol z_i^{0})_{[1]},c_1,\cdots, c_M]^\top).
\end{equation}

In the graph domain, node curvature $Ric(i)$ is defined by averaging the Ricci curvature in its neighborhood, in analogy to tracing around the tangent space of the manifold.
That is, the node-level curvature on the graph is formulated as 
$Ric(i)=\frac{1}{degree_i}\sum\nolimits_{j \in \mathcal N_i} Ric(i,j)$, 
where $degree_i$ is the degree of $v_i$ and $\mathcal N_i$ denotes the $1$-hop neighborhood of node $i$.
Then, we propose a node-level curvature consistency loss as
  \vspace{-0.05in}
\begin{equation}
\mathcal L_{Curv}=\frac{1}{N}\sum\nolimits_i |Ric(i)-\bar{c}_i|^2,
\end{equation}
so that \emph{curvatures of factor manifolds are jointly learnt with the model via the fine-grained curvature modeling.}

Till now, we construct the heterogeneous curvature space modeling the fine-grained curvatures of the graph.
Thereby, \emph{the constructed curvature space supports geometric graph clustering with the Ricci loss, which requires a closer look over the various Ricci curvatures on the graph} (Eqs. \ref{intra}-\ref{ricciLoss}).


\noindent \textbf{Remarks.} 
The advantages of our design are
1) $\mathcal P_H$ supports node-level curvature modeling for geometric clustering, and its factors has learnable curvatures, different from the product manifolds in \citet{GuSGR19,DBLP:conf/www/WangWSWNAXYC21}.
2) $\mathcal P_H$ as a whole owns the closed form expression of geometric metrics inherited from its factor manifolds.
3) $\mathcal P_H$ decomposes itself into $(M+1)$ \emph{different geometric views} corresponding to each factor (i.e., $M$ \emph{restricted views} and $1$ \emph{free view}).

\vspace{-0.03in}
\subsubsection{Generate Deep Representations in the Product Manifold}

Thanks to the product construction, encoding in the heterogeneous curvature space is transformed into encoding in each factor manifold.
Most of the Riemannian GCNs involve the tangent space out of the original manifold, and recent studies observe the inferior of tangential methods \citep{DBLP:conf/cvpr/DaiWGJ21}. 

To bridge this gap, we design a \emph{fully Riemannian} GCN (\emph{\textbf{f}RGCN}) for the restricted factor $\mathcal M^{c, d}$, whose novelty lie in that \emph{all the operations are fully Riemannian for any $c$}, i.e., 
no tangent space is involved. We design the manifold preserving operators of \emph{\textbf{f}RGCN} as follows.

\noindent \textbf{Feature Transformation.} 
First, we formulate a generalized Lorentz Transformation ($gLT$) for dimension transformation, inspired by the classic LT.
The transform $\mathcal M^{c, d_m} \to \mathcal M^{c, d_n}$ is done via the matrix left-multiplication with the transform matrix derived as follows,
\vspace{-0.07in}
\begin{equation}
gLT_{\boldsymbol z}^{c, d_m \to d_n}(\boldsymbol W)=
\left[\begin{array}{cc}
w_t & \mathbf{0}^{\top} \\
\mathbf{0} &  \boldsymbol W
\end{array}\right].
\end{equation}
Recall that $\boldsymbol z=[z_t \  \boldsymbol z_s]^\top \in \mathcal M^{c, d_m}$.
In $gLT$, $w_t$ is responsible to scale $z_t$ while $\boldsymbol W$ transforms  $\boldsymbol z_s$. 
We derive the closed-form $t$-scaling as $w_t= \frac{1}{z_t}\sqrt{sgn(c)\left(\frac{1}{c}-\ell( \boldsymbol W, \boldsymbol z_s ) \right)}$ and $\ell( \boldsymbol W, \boldsymbol z_s )=\left\| \boldsymbol W \boldsymbol z_s\right\|^2$.

Now, we prove that the transformed feature with $gLT$ resides in the target manifold.
\newtheorem*{prop2}{Proposition 2 (Manifold Preserving)} 
\begin{prop2}
$\forall \boldsymbol z \in \mathcal M^{c, d_m}, \forall c$, 
 $gLT^{c, d_m \to d_n}_{\boldsymbol z}(\boldsymbol W) \boldsymbol z \in \mathcal M^{c, d_n}$ holds for any $\boldsymbol W \in \mathbb R^{d_n \times d_m}$.
\end{prop2}
\vspace{-0.07in}
\begin{proof}
\vspace{-0.05in}
Please refer to Appendix.
\end{proof}
\vspace{-0.05in}
\noindent Note that, the classic LT works with a fixed dimension. Recently, \cite{DBLP:conf/cvpr/DaiWGJ21} optimize with orthogonal constraint unfriendly to deep learning. \cite{DBLP:conf/acl/ChenHLZLLSZ22} restrict in negative curvature. That is, all of them cannot satisfy our need.

Second, we add the bias for $gLT$ and obtain the linear layer in the manifold of any curvature $c$ as follows,
\vspace{-0.1in}
\begin{equation}
LL^c(\boldsymbol W,\boldsymbol z, \boldsymbol b)=
\left[\begin{array}{c}
w_t z_t\\
 \boldsymbol W \boldsymbol z_s+\boldsymbol b
\end{array}\right],
\label{ll}
\end{equation}
where $\boldsymbol b$ is the bias and $\ell( \boldsymbol W, \boldsymbol z_s )=\left\| \boldsymbol W\boldsymbol z_s +\boldsymbol b\right\|^2$. It is easy to check that $LL^c$ is \emph{manifold preserving}.

\noindent \textbf{Attentive Aggregation.}
The encoding of $i$ is updated as the weighted geometric centorid over the set $\bar{\mathcal N}_i$, the neighbors of $i$ and itself, i.e.,
$\arg \min\nolimits_{\boldsymbol h_i \in \mathcal M} \sum\nolimits_{j\in \bar{\mathcal N}_i} \nu_{ij} d^2_c(\boldsymbol h_i, \boldsymbol h_j), \forall c$ and $\nu_{ij}$ denotes the attentive weight.
For any $c$, we derived the closed form solution $\boldsymbol h_i=AGG^c(\{\boldsymbol h_j, \nu_{ij}\}| j \in \bar{\mathcal N}_i)$,
\vspace{-0.07in}
\begin{equation}
\resizebox{0.96\hsize}{!}{$
AGG^c(\{\boldsymbol h_j, \nu_{ij}\} | j \in \bar{\mathcal N}_i )= 
\frac{1}{\sqrt{|c|} } \sum\nolimits_{j \in \bar{\mathcal N}_i}\frac{\nu_{ij} \boldsymbol h_j}{\left| \|\sum\nolimits_{j \in \bar{\mathcal N}_i} \nu_{ij} \boldsymbol h_j\|_c \right|}.
$}
\label{agg}
\end{equation}
The attentive weights $\nu_{ij}$ is the importance of $j$ in the aggregation over $\bar{\mathcal N}_i$.  
We define the attentive weights  based on the distance  in the manifold, $\nu_{ij}=Softmax(-\tau d^c(\boldsymbol h_j, \boldsymbol h_i)-\gamma)$,
where $\tau$ is an inverse temperature and we add a bias $\gamma$.
It is easy to check that the centroid in Eq. (\ref{agg}) lives in the manifold, $\forall c$, and thus $AGG^c$ is \emph{manifold preserving}.

Note that, Einstein midpoint formulates an arithmetic mean in the manifold but lacks geometric interpretation. 
Fr\'{e}chet mean elegantly generalizes from Einstein midpoint but does not offer any closed form solution \citep{DBLP:conf/acl/ChenHLZLLSZ22}.
Our closed form solution in Eq. (\ref{agg}), generalizing to any curvature, is the geometric centroid of squared distance. 

 \emph{The Free Factor.} Linear layer $LL^0$ is done via replacing $LL^c$ with a free $w_t \in \mathbb R$.
Attentive aggregation is defined as $AGG^0(\{\boldsymbol h_j, \nu_{ij}\}| j \in \bar{\mathcal N}_i)=\sum_{j \in \bar{\mathcal N}_i} \nu_{ij}\boldsymbol h_j$ where attentive weights $\nu_{ij}$ is computed based on distance $d_0$. 
They are manifold preserving as there is no norm restriction in $\mathbb R^{d_0}$.

\vspace{-0.07in}
\subsection{Learning by Reweighted Geometric Contrasting} \label{3}
\vspace{-0.02in}
In this subsection, we train the graph clusters with a contrastive loss in the proposed curvature space.
Specifically, we propose a Reweighted Geometric Contrasting (RGC) approach,
in which we contrast across different geometric views with a novel dual reweighting, as shown in Fig 1 (c).

\vspace{-0.05in}
\subsubsection{Augmentation-Free Geometric Contrast}
The augmentation is nontrivial for graph contrastive learning, and requires special design for clustering \citep{DBLP:conf/ijcai/GongZTL22}. 
Instead, our \textsc{Congregate} is free of augmentation where we take advantage of the carefully designed $\mathcal P_H$ for  contrastive learning. 
Thanks to the product construction, $\mathcal P_H$ itself owns different \emph{geometric views} as remarked in Sec. \ref{2}.
The contrast strategy is that we contrast each restricted view in $\mathcal M_m^{c_m, d_m}$ with the free view in $\mathbb R^{d_0}$, and vice versa.

The remaining challenge is how to contrast between different manifolds, i.e., $\mathcal M_m^{c_m, d_m}$ and $\mathbb R^{d_0}$.
The difference in both curvature and dimension blocks the application of typical similarity functions.
We propose to bridge this gap by $gLT$ and bijection $\psi_{\mathcal M \to \mathbb R}$ of \emph{Diffeomorphism}.
(Recall that we have already provided an effective mathematics tool for dimension transformation, $gLT$.)
Specifically, we introduce an image of  restricted  view $\hat{\boldsymbol z}^{m}$ that is comparable with the free view.
First, we employ $gLT$ to transform $\boldsymbol z^m$ into $\mathcal M_m^{c_m, d_0-1}$ whose ambient space is $\mathbb R^{d_0}$.
Second, we apply the diffeomorphism bijection and thus the image is given as follows,
\vspace{-0.09in}
\begin{equation}
\hat{\boldsymbol z}^{m}=\psi_{\mathcal M \to \mathbb R}(gLT_{\boldsymbol z^{m}}^{c_m, d_m \to (d_0-1)}(\mathbf W)\boldsymbol z^{m}),
\vspace{-0.01in}
\label{m2e}
\end{equation}
where  parameter $\mathbf W$ characterizes $gLT$, and $log^{c_m}_\mathbf 0( \cdot)$ is utilized as the bijection since its differentiable inverse exists $log^{c_m}_\mathbf 0(exp^{c_m}_\mathbf 0(\boldsymbol z))=\boldsymbol z$.
Note that $\hat{\boldsymbol z}^{m}_i \in \mathbb R^{d_0}$.
Then, we define the similarity as a bilinear critic with parameter $\mathbf S$,
\vspace{-0.02in}
\begin{equation}
Sim(\boldsymbol z^{m}, \boldsymbol z^{0}) = (  {\hat{\boldsymbol z}^{m}} )^\top  \mathbf S \boldsymbol z^{0}.
\label{sim}
\vspace{-0.01in}
\end{equation}
Our formulation of Eq. (\ref{sim}) does not introduce additional tangent space, and its advantage is examined in Sec. \ref{4-3}.

\vspace{-0.05in}
\subsubsection{Dual Reweighting in Curvature Space}
A drawback of the popular InfoNCE loss is hardness unawareness (equally treating the hard sample pairs and the easy ones), limiting the discriminative ability \citep{DBLP:conf/iclr/RobinsonCSJ21}.
To address this issue, we propose a \emph{dual reweighting}, paying more attention to both hard negatives and hard positives for contrastive learning in curvature space.


First, we specify the hard samples in the context of graph clustering where cluster assignment offers pseudo labels.
Intuitively, the nodes assigned to different clusters but sharing large similarity are referred to as \emph{hard negatives},
while 
\emph{the border nodes sharing small similarity to the cluster centroid are hard positives}.
Second, we model the hardness by comparing cluster assignment (pseudo label) and representation similarity, and formulate the dual reweighting as follows,
\vspace{-0.05in}
\begin{equation}
\mathcal W(\boldsymbol z^{m}_i, \boldsymbol z^{0}_j)=|{\boldsymbol \pi_i}^\top\boldsymbol \pi_j - Sim(\hat{\boldsymbol z}^{m}_i, \boldsymbol z^{0}_j)|^\beta
\label{reweight}
\end{equation}
where the control coefficient $\beta$ is a positive integer, and $\mathcal W(\boldsymbol z^{m}_i, \boldsymbol z^{0}_j)$ up-weights both hard positives and hard negatives while down-weighting the easy ones.

Recently, 
\cite{DBLP:conf/cikm/0008YPY22} design a Riemannian reweighing for node embedding only and thus fail to consider clusters.
\cite{DBLP:journals/corr/abs-2212-08665} select hard positives in Euclidean space while we need to handle different manifolds. 
Both of them cannot meet our need and motivate our design of Eq. (\ref{reweight}).


\begin{algorithm}[t]
        \caption{\textbf{ Training \textsc{Congregate}  }       } 
        \KwIn{Graph $G$, \#(Clusters)$=$\emph{K}, \#(Factors)$=$(\emph{M+1})}
        \KwOut{Encoder $f$, Cluster centroids $\{\boldsymbol \phi_k\}_{k=1, \cdots, K}$}
        Preprocessing: Compute Ricci curvatures on $G$;\\
\While{not converging}{   
            Create geometric views $[\boldsymbol z^0 \ \boldsymbol z^1 \cdots \boldsymbol z^M] \gets$ \emph{\textbf{f}RGCN};\\
            \For{each restricted view $\boldsymbol z^m, m \in [1, M]$ }{
              /*  \textit{Contrast with the free view   $\boldsymbol z^0$}  */\\
                Node-to-Node contrast based on Eq. (\ref{n2n});\\
                Node-to-Cluster contrast based on Eq. (\ref{n2c});\\
            }
            Train  $\{\boldsymbol \phi_k\}_{k=1, \cdots, K}$ by optimizing $\mathcal J$ in Eq. (\ref{all});\\
}
\end{algorithm}

\vspace{-0.05in}
\subsubsection{Node-to-Node \& Node-to-Cluster Contrasting}
The RGC loss consists of Node-to-Node and Node-to-Cluster contrasting, 
where we contrast different geometric views with the dual reweighting and $Sim$ function in generic curvature space.
First, we define Node-to-Node contrast loss as follows,
\vspace{-0.18in}
\begin{equation}
I(\mathbf Z^m, \mathbf Z^0)=
- \sum\nolimits_{i=1}^N log \frac{ e^{\mathcal W(\boldsymbol z^{m}_i, \boldsymbol z^{0}_i)Sim(\boldsymbol z^{m}_i, \boldsymbol z^{0}_i)} }
{\sum\nolimits_{j=1}^N e^{\mathcal W(\boldsymbol z^{m}_i, \boldsymbol z^{0}_j)Sim(\boldsymbol z^{m}_i, \boldsymbol z^{0}_j)}}.
\vspace{-0.02in}
\label{n2n}
\end{equation}

\noindent Second, we contrast node encoding of one view with cluster centroids of another view, and formulate the Node-to-Cluster contrast loss as follows, 
\vspace{-0.12in}
\begin{equation}
I(\mathbf Z^m, \mathbf \Phi^0)=
- \sum\nolimits_{i =1}^N log \frac{ e^{\mathcal W(\boldsymbol z^{m}_i, \boldsymbol \phi^{0}_{k_i})Sim(\boldsymbol z^{m}_i, \boldsymbol \phi^{0}_{k_i})} }
{\sum\nolimits_{k=1}^K e^{\mathcal W(\boldsymbol z^{m}_i, \boldsymbol \phi^{0}_{k})Sim(\boldsymbol z^{m}_i, \boldsymbol \phi^{0}_k)}},
\label{n2c}
\vspace{-0.02in}
\end{equation}
where node $v_i$ is assigned to cluster $k_i$. Here, in $\mathcal W(\boldsymbol z^{m}_i, \boldsymbol \phi^{0}_{k_i})$, the inner product term is simplified as $[\boldsymbol \pi_i]_{k_i}$ the probability of $v_i$ assigned to cluster $k_i$.
Thus, we have RGC loss as follows,
\vspace{-0.22in}
\begin{equation}
\mathcal L_{RGC}=\sum\nolimits_{m=1}^M \sum\nolimits_{\mathbf X \in \{\mathbf Z^0, \mathbf \Phi^0\}}( I(\mathbf Z^m, \mathbf X)  +I(\mathbf X, \mathbf Z^m) ).  
\label{contrast}
\end{equation}

\noindent In our curvature space, intra-cluster node similarity is maximized as they positively contrast to the same centroid, while inter-cluster nodes are separated by negative contrast.
Meanwhile, more attention is paid to the similar cluster centorids (\emph{hard negatives}) and the nodes residing in the cluster border (\emph{hard positives}), thanks to dual reweighting of Eq. (\ref{reweight}).

\noindent \textbf{The Overall Loss} of our model is finally defined as follows,
\vspace{-0.03in}
\begin{equation}
\mathcal J = \mathcal L_{Ric} + \alpha_1 \mathcal L_{Curv}+ \alpha_2 \mathcal L_{RGC}, 
\label{all}
\end{equation}
where $\alpha_1$ and $\alpha_2$ are weighting coefficients. We summarize the training process in Algo.  1.
In this way, we end-to-end train the cluster centorids in the proposed curvature space.

\begin{table*}
\vspace{-0.1in}
\centering
\resizebox{1.05\linewidth}{!}{ 
\begin{tabular}{c  c | c c  c |  c c c | c c c| c c c  }
\toprule
&  \multirow{2}{*}{\textbf{Method}}  & \multicolumn{3}{c|}{ \textbf{Cora}} & \multicolumn{3}{c|}{\textbf{ Citeseer} } & \multicolumn{3}{c|}{\textbf{ MAG-CS} }   & \multicolumn{3}{c}{\textbf{ Amazon-Photo} } \\
&     & ACC & NMI & ARI  & ACC & NMI & ARI  & ACC & NMI & ARI   & ACC & NMI & ARI  \\
\toprule
\multirow{5}{*}{\rotatebox{90}{    \emph{SS}  } } 
&GAE  \citep{kipf2016variational}         
&   61.3 \tiny{(0.8)}    & 44.4 \tiny{(1.1)}    & 38.1 \tiny{(0.9)}     &   61.4 \tiny{(0.8)}     &  34.6 \tiny{(0.7)}     & 33.6 \tiny{(1.2)} 
&   63.2 \tiny{(2.6)}    & 69.9 \tiny{(0.6)}    & 52.8 \tiny{(1.5)}     &   71.6 \tiny{(2.5)}     &  62.1 \tiny{(2.8)}     &  48.8 \tiny{(4.6)}     \\
&VGAE   \citep{kipf2016variational}        
&   64.7 \tiny{(1.3)}    & 43.4 \tiny{(1.6)}    & 37.5 \tiny{(2.1)}     &   61.0 \tiny{(0.4)}     &  32.7 \tiny{(0.3)}    & 33.1 \tiny{(0.5)} 
&   60.4 \tiny{(2.9)}    & 65.3 \tiny{(1.4)}    & 50.0 \tiny{(2.1)}     &  74.3 \tiny{(3.6)}     &  66.0 \tiny{(3.4)}      & 56.2 \tiny{(4.7)}     \\
&DGI    \citep{VelickovicFHLBH19}                                    
&   72.6 \tiny{(0.9)}    & 57.1 \tiny{(1.7)}    & 51.1 \tiny{(3.0)}     &   68.6 \tiny{(1.3)}     &  43.5 \tiny{(1.2)}   & 44.5 \tiny{(1.9)} 
&   60.0 \tiny{(0.6)}    & 65.9 \tiny{(0.4)}    & 50.3 \tiny{(0.9)}     &   57.2 \tiny{(1.9)}     &  37.6 \tiny{(0.3)}      & 26.4 \tiny{(0.3)}     \\
&ARGA   \citep{DBLP:journals/tcyb/PanHFLJZ20}    
&   71.0 \tiny{(2.5)}    & 51.1 \tiny{(0.5)}    & 47.7 \tiny{(0.3)}     &   61.1 \tiny{(0.5)}     &  34.4 \tiny{(0.7)}   & 33.4 \tiny{(1.2)} 
&   47.9 \tiny{(6.0)}    & 48.7 \tiny{(3.0)}    & 23.6 \tiny{(9.0)}     &   69.3 \tiny{(2.3)}     &  58.4 \tiny{(2.8)}   & 44.2 \tiny{(4.4)}  \\
&MVGRL \citep{HassaniA20}          
&  70.5 \tiny{(3.7)}    & 55.6 \tiny{(1.5)}    & 48.7 \tiny{(3.9)}     &   62.8 \tiny{(1.6)}     & 40.7 \tiny{(0.9)}   & 34.2 \tiny{(1.7)} 
&  61.6 \tiny{(3.3)}    & 65.4 \tiny{(1.9)}    & 49.2 \tiny{(5.3)}     &  41.1 \tiny{(3.2)}     &  30.3 \tiny{(3.9)}      &  18.8 \tiny{(2.3)}     \\
\midrule
\multirow{13}{*}{\rotatebox{90}{    \emph{Deep  Clustering}   } } 
&DAEGC    \citep{DBLP:conf/ijcai/WangPHLJZ19}
&   70.4 \tiny{(0.4)}    & 52.9 \tiny{(0.7)}    & 49.6 \tiny{(0.4)}           &   64.5 \tiny{(1.4)}     &  36.4 \tiny{(0.9)}   & 37.8 \tiny{(1.2)} 
&   48.1 \tiny{(3.8)}    & 60.3 \tiny{(0.8)}    & 47.4 \tiny{(4.2)}           &   76.0 \tiny{(0.2)}     &  65.3 \tiny{(0.5)}   & 58.1 \tiny{(0.2)}  \\
&SDCN   \citep{DBLP:conf/www/Bo0SZL020}
&   35.6 \tiny{(2.8)}    & 14.3 \tiny{(1.9)}    & $\ \ $7.8 \tiny{(3.2)}  &   65.9 \tiny{(0.3)}     &  38.7 \tiny{(0.3)}   & 40.2 \tiny{(0.4)} 
&   51.6 \tiny{(5.5)}    & 58.0 \tiny{(1.9)}    & 46.9 \tiny{(8.1)}        &   53.4 \tiny{(0.8)}     &  44.9 \tiny{(0.8)}   & 31.2 \tiny{(1.2)}  \\
&AGE    \citep{DBLP:conf/kdd/CuiZY020}                     
&   73.5 \tiny{(1.8)}    & 57.6 \tiny{(1.4)}    & 50.1 \tiny{(2.1)}     &  69.7 \tiny{(0.2)}     & 44.9 \tiny{(0.5)}   & 45.3 \tiny{(0.4)} 
&   59.1 \tiny{(1.7)}    & 66.7 \tiny{(0.3)}    & 51.1 \tiny{(2.8)}     &   75.9 \tiny{(0.7)}     &  65.4 \tiny{(0.6)}   & 55.9 \tiny{(1.3)}  \\
&GMM-VGAE      \citep{DBLP:conf/aaai/HuiZH20}        
&   71.5 \tiny{(0.2)}    & 53.1 \tiny{(2.1)}    & 47.4 \tiny{(0.6)}     &  67.5 \tiny{(0.9)}     & 40.7 \tiny{(1.1)}   & 42.4 \tiny{(0.5)} 
&   67.2 \tiny{(2.6)}    & 72.8 \tiny{(0.7)}    & 56.1 \tiny{(1.9)}     &  75.5 \tiny{(0.3)}     & 68.1 \tiny{(0.7)}   & 57.9 \tiny{(0.9)}  \\
&AGCN    \citep{DBLP:conf/mm/PengLJH21}          
&   72.2 \tiny{(3.6)}    & 54.7 \tiny{(1.3)}    & 48.9 \tiny{(2.7)}     &   68.8 \tiny{(0.2)}     &  41.5 \tiny{(0.3)}   & 43.8 \tiny{(0.3)} 
&   54.2 \tiny{(5.2)}    & 59.4 \tiny{(2.1)}    & 49.2 \tiny{(6.5)}     &   45.2 \tiny{(1.0)}     &  41.6 \tiny{(1.1)}   & 36.6 \tiny{(0.2)}  \\
&GDCL   \citep{DBLP:conf/ijcai/ZhaoYWYD21}                    
&   70.8 \tiny{(0.5)}    & 56.6 \tiny{(0.4)}    & 48.1 \tiny{(0.7)}     &  66.4 \tiny{(0.6)}     & 39.5 \tiny{(0.4)}   & 41.1 \tiny{(1.0)} 
&   53.9 \tiny{(3.1)}    & 60.3 \tiny{(0.8)}    & 48.8 \tiny{(5.1)}     &  43.8 \tiny{(0.8)}     &  37.3 \tiny{(0.3)}   & 21.6 \tiny{(0.5)}  \\
&S$^3$GC   \citep{S3GC}                       
&   74.2 \tiny{(3.0)}    & 58.9 \tiny{(1.8)}    & 54.4 \tiny{(2.5)}     &  68.8 \tiny{(1.7)}     & 44.1 \tiny{(0.9)}   & 44.8 \tiny{(0.6)} 
&   65.4 \tiny{(2.3)}    & 77.6 \tiny{(0.6)}    & 61.9 \tiny{(2.4)}     &  71.8 \tiny{(0.2)}     & 63.7 \tiny{(0.8)}   & 45.8 \tiny{(1.0)}  \\
&CGC   \citep{DBLP:journals/corr/abs-2204-08504}                     
&   73.1 \tiny{(2.2)}    & 57.0 \tiny{(0.9)}    & 49.3 \tiny{(1.8)}     &   69.6 \tiny{(0.6)}     &  44.6 \tiny{(0.1)}   & 46.0 \tiny{(0.6)} 
&   69.3 \tiny{(4.0)}    & \underline{79.3} \tiny{(1.2)}  & 64.4 \tiny{(3.7)} &  75.2 \tiny{(0.1)}  & 64.1 \tiny{(1.2)}   & 51.7 \tiny{(0.6)}  \\
&gCooL  \citep{DBLP:conf/www/LiJT22}             
&   72.0 \tiny{(1.6)}    & 58.3 \tiny{(0.2)}  & 56.9 \tiny{(0.9)}    & \underline{71.5} \tiny{(1.0)}    & 47.3 \tiny{(0.8)}    & 46.8 \tiny{(1.5)} 
&   70.7 \tiny{(1.5)}    & 78.6 \tiny{(1.0)}  & 66.0 \tiny{(1.6)}     &  72.7 \tiny{(0.6)}    & 63.2 \tiny{(0.0)}      & 52.4 \tiny{(0.2)}     \\
&HostPool  \citep{DBLP:conf/cikm/DuvalM22}            
&   71.8 \tiny{(0.8)}  & 60.5 \tiny{(1.0)}  & \underline{58.5} \tiny{(1.1)}  & 70.9 \tiny{(0.5)}   & \underline{50.2} \tiny{(0.3)} &45.9 \tiny{(0.7)} 
&   67.5 \tiny{(2.1)}   & 79.0 \tiny{(0.6)}    & 67.1 \tiny{(1.2)}     &  69.4 \tiny{(0.1)}     &  59.8 \tiny{(1.0)}   & 45.5 \tiny{(0.7)} \\
&AGC-DRR \citep{DBLP:conf/ijcai/GongZTL22}  
&  40.6 \tiny{(0.6)}    & 18.7 \tiny{(0.7)}    & 14.8 \tiny{(1.6)}     &  68.3 \tiny{(1.8)}     & 43.3 \tiny{(1.4)}   & 45.3 \tiny{(2.3)} 
&  71.2 \tiny{(0.8)}    & 71.2 \tiny{(0.8)}    & 65.8 \tiny{(3.1)}     &  76.8 \tiny{(1.5)}     & 66.5 \tiny{(1.2)}   & \underline{60.1} \tiny{(1.6)}  \\
&FT-VGAE  \citep{DBLP:conf/ijcai/MrabahBK22}
&  \underline{77.4} \tiny{(1.1)}  & \underline{61.0} \tiny{(0.5)} &58.2 \tiny{(1.3)} & 70.8 \tiny{(0.5)} &44.5 \tiny{(0.1)}&46.7 \tiny{(0.7)} 
&  \underline{73.3} \tiny{(1.0)} & 69.5 \tiny{(0.5)} &67.9 \tiny{(2.2)}    & \underline{78.1} \tiny{(1.0)}  & \underline{69.8} \tiny{(0.7)}   & 59.3 \tiny{(0.8)}  \\
&HSAN   \citep{DBLP:journals/corr/abs-2212-08665} 
&   77.1 \tiny{(1.6)}    & 59.2 \tiny{(1.0)}    & 57.5 \tiny{(2.7)}     &  71.2 \tiny{(0.8)}     & 45.1 \tiny{(0.7)}   & \underline{47.1} \tiny{(1.1)} 
&   72.8 \tiny{(1.0)}    & 77.3 \tiny{(0.9)}    & \underline{68.2} \tiny{(1.7)}     &  77.0 \tiny{(0.3)}     &  67.2 \tiny{(0.3)}   & 58.0 \tiny{(0.5)}  \\
\midrule
\multirow{2}{*}{\rotatebox{90}{    \emph{R}   } } 
&  RiccCom  \citep{CDRiccFlow} 
&  55.6 \tiny{(0.3)}    & 58.1 \tiny{(1.1)}  & 48.9 \tiny{(0.9)}     &  67.3 \tiny{(1.2)}     & 46.2 \tiny{(0.8)}   & 44.9 \tiny{(0.6)} 
&  69.1 \tiny{(3.2)}    & 71.6 \tiny{(1.2)}  & 65.2 \tiny{(0.8)}     &   58.3 \tiny{(2.3)}    & 62.9 \tiny{(0.9)}   & 57.4 \tiny{(0.9)}  \\
& \textbf{\textsc{Congregate} (Ours)}  
&  \textbf{78.5} \tiny{(1.0)}  & \textbf{63.2} \tiny{(0.5)}  &\textbf{59.3} \tiny{(1.2)} &\textbf{72.7} \tiny{(0.6)} &\textbf{50.9} \tiny{(0.3)}   & \textbf{48.3} \tiny{(1.0)} 
&  \textbf{73.3} \tiny{(0.7)}  & \textbf{80.8} \tiny{(1.3)} & \textbf{69.5} \tiny{(1.1)} &\textbf{79.7} \tiny{(1.8)} &\textbf{71.3} \tiny{(0.5)}   & \textbf{60.9} \tiny{(1.1)}  \\
\bottomrule
\end{tabular} 
}
\vspace{-0.05in}
\caption{Clustering results of 20 methods on Cora, Citerseer, MAG-CS  and Amazon-Photo in terms of NMI, ARI and ACC (\%). Standard derivation is given in brackets. The best results are highlighted in \textbf{bold}, and the runner up \underline{underlined}. }
\vspace{-0.1in}
\label{results}
\end{table*}


\noindent \textbf{Complexity Analysis.}
Eq. (\ref{contrast}) is the most costly, yielding the computational complexity of $O(2M|\mathcal V|^2+2MK|\mathcal V|)$. It is similar to typical contrastive methods \citep{VelickovicFHLBH19,HassaniA20}.
The Ricci curvatures only need to be computed once as a pre-processing, and can be effectively obtained similar to \cite{CDRiccFlow,DBLP:conf/iclr/YeLM0020}.

\vspace{-0.05in}
\section{Experiment}
In this section, we evaluate our model with 20 baselines on 4 public datasets, aiming to answer the following research questions (\emph{RQs}),
\begin{itemize}
  \item[-] \textbf{\emph{RQ1}}: How does the proposed \textsc{Congregate} perform?
  \item[-] \textbf{\emph{RQ2}}: What are the effects of the proposed components?
  \item[-] \textbf{\emph{RQ3}}: Why does \emph{Ricci Curvature} work?
\end{itemize}




\vspace{-0.05in}
\subsection{Experimental Setups}

\noindent\textbf{Datasets \& Baselines.}
We choose $4$ datasets, i.e.,
Cora and Citeseer \citep{S3GC}, and larger MAG-CS  \citep{DBLP:journals/corr/abs-2204-08504} and Amazon-Photo \citep{DBLP:conf/www/LiJT22}. 
We focus on deep graph clustering with no labels available.
Thus, both the strong deep clustering methods (\emph{DC}) and self-supervised learning methods  (\emph{SS}) are included as \emph{Euclidean Baselines} for a comprehensive evaluation.
There are $13$ strong DC methods and $5$ SS methods, summarized in Table \ref{results}.
There exists few \emph{Riemannian Baselines} (\emph{R}). 
Note that, recent Riemannian GNNs do not have clustering ability, as typical clustering algorithms cannot be directly applied/incorporated owing to the inherent difference in geometry.
Instead, we choose a recent shallow model, \emph{RicciCom}.
We are the first to bridge Riemannian space and graph clustering to our knowledge.

\noindent\textbf{Evaluation Protocol.}
We employ $3$ popular evaluation metrics, i.e., Normalized Mutual Information (NMI), Adjusted Rand Index (ARI) and Accuracy (ACC) \citep{S3GC,DBLP:conf/www/LiJT22,DBLP:conf/ijcai/MrabahBK22}.
The number of clusters $K$ is set as the number of real classes on each dataset. 
We perform $10$ independent runs, and report the mean value with standard deviation for fair comparisons.
For the encoding-clustering baselines, we apply $K$-means to obtain the results.

\noindent\textbf{Reproducibility.}
Further details and code are provided https://github.com/CurvCluster/Congregate.
If input features live in the Euclidean space, we use the inverse bijection $\psi^{-1}_{\mathcal M \to \mathbb R}$ in Eq. (\ref{m2e}) to map the Euclidean input to a factor manifold.
In \emph{\textbf{f}RGCN}, the convolutional layer is stacked twice. Parameters living in the factor are optimized via Riemannian Adam \citep{DBLP:journals/corr/abs-2005-02819}. 
We utilize a $2$-layer MLP to approximate the fine-grained curvature. In RGC loss, hyperparameter $\beta$ of the reweighting  is $2$ as default.

\noindent(\emph{Details on datasets, codes and proofs are in the Appendix.})

\vspace{-0.1in}
\subsection{Empirical Results} \label{4-3}

\noindent\textbf{\emph{RQ1}: Main Results.}
The clustering results on all the datasets in terms of NMI, ARI and ACC are reported in Table \ref{results}.
Our \textsc{Congregate} is instantiated with $4$  factor manifolds whose dimensionality are $\{32, 32, 16, 16\}$, and it consistently achieve the best results among $19$ competitors.
The reasons are 1) we take advantage of the proposed curvature space and the consensus clustering from different geometric views.
2) We  jointly learns high discriminative node encodings and cluster centroids with the proposed reweighting loss. 

\noindent\textbf{\emph{RQ2}: Ablation Study.}
We investigate on how each proposed component contributes to the success of our \textsc{Congregate}:
i)  \emph{\textbf{f}RGCN} for modeling graph fully Riemannianly, ii) $\varphi \circ gLT$ for contrasting between different manifolds and iii) the dual reweighting in $\mathcal W$ for paying attention to hard samples.

To evaluate the effectiveness of \emph{\textbf{f}RGCN}, we introduce a variant which replaces \emph{\textbf{f}RGCN} with a $GAT^c$. Concretely, $GAT^c$ generalizes GAT \citep{velickovic2018graph} in a manifold of curvature $c$ with tangent spaces.
We utilize the tangential methods of any $c$ formulated in \cite{SkopekGB20}.
To evaluate the effectiveness of $\varphi \circ gLT$, we introduce a variant using $\mathbf T log^{c_m}_\mathbf 0$ instead, where the matrix $\mathbf T$ is given for dimension transformation. It introduce an additional tangent space compare to the design in our model.
To evaluate the effectiveness of $\mathcal W$,  we introduce two kinds of variants. The first variant (denoted as $-pAware$) removes the $\mathcal W$ on the numerators of our RGC loss, thus keeping the attention to hard negatives only. 
The second variant (denoted as $-hard$) eliminates all the $\mathcal W$, resulting in an InfoNCE in Riemannian space without hardness awareness. 
In addition, we examine the effect on the number of factor manifolds. To this end, the variants above are instantiated in product space of $4$ factors and $5$ factors, respectively.
We report the NMI and ACC of the clustering results on Cora and Citeseer datasets in Table 2, and find that: 
\textbf{i)} Our \textsc{Congregate} beats the $-$\emph{\textbf{f}RGCN} and $-\varphi \circ gLT$. 
It shows that introducing addtional tangent spaces trends to results in inferior clustering, and thus \emph{testifies the effectiveness of fully Riemannian model.}
\textbf{ii)}  The product space $5$ factors outperforms that of $4$ factors. It suggests that more factor manifolds may benefit the performance, and the reason is that more factors give further flexibility for the fine-grained curvature modeling.
\textbf{iii)}  $-posA$ variant performs better than $-hard$, and the proposed RGC loss is the best. 
It shows the importance of hard samples, and \emph{more attentions to hard positives (the border nodes) further help the performance}, which is the reason of our design that we pay more attentions to both hard positives and hard negatives.

\begin{table}[t]
\centering
\resizebox{0.95\linewidth}{!}{ 
\begin{tabular}{c  l | c c | c c}
\toprule
 &  \multirow{2}{*}{\textbf{Variant}}  & \multicolumn{2}{c|}{ \textbf{Cora}} & \multicolumn{2}{c}{\textbf{ Citeseer} }   \\
     &     & ACC & NMI  & ACC & NMI     \\
\toprule
\multirow{5}{*}{\rotatebox{90}{   4 factors } } 
&\textbf{\textsc{Congregate}}  & \textbf{78.5} \tiny{(1.0)} & \textbf{63.2} \tiny{(0.5)} & \textbf{72.7} \tiny{(0.6)} & \textbf{50.9} \tiny{(0.3)} \\
&$-$\emph{f}RGCN    & 75.9 \tiny{(0.5)} & 60.4 \tiny{(0.2)} & 72.0 \tiny{(0.9)} & 48.3 \tiny{(0.6)} \\
&$-\varphi \circ gLT$   & 77.5 \tiny{(0.8)} & 61.7 \tiny{(0.6)} & 71.9 \tiny{(1.3)} & 47.9 \tiny{(0.2)} \\
&$-pAware$                 &  77.2 \tiny{(0.7)} & 62.1 \tiny{(0.8)} & 70.3 \tiny{(0.5)} & 49.0 \tiny{(1.4)} \\
&$-hard$                       & 76.8 \tiny{(1.1)} & 61.5 \tiny{(0.3)} & 69.8 \tiny{(0.2)} & 48.9 \tiny{(0.5)}  \\
\midrule
\multirow{5}{*}{\rotatebox{90}{   5 factors  } } 
&\textbf{\textsc{Congregate}}  & \textbf{78.1} \tiny{(0.9)} & \textbf{63.8} \tiny{(0.4)} & \textbf{73.1} \tiny{(0.6)} & \textbf{52.4} \tiny{(0.7)} \\
&$-$\emph{f}RGCN    &  76.3 \tiny{(1.2)}  & 61.9 \tiny{(0.5)} & 72.3 \tiny{(1.0)} & 49.8 \tiny{(0.9)} \\
&$-\varphi \circ gLT$   &  77.8 \tiny{(0.6)} & 62.5 \tiny{(0.9)} & 72.8 \tiny{(0.7)} & 51.6 \tiny{(0.4)} \\
&$-pAware$                  &  77.3 \tiny{(2.2)} & 63.0 \tiny{(0.3)} & 71.2 \tiny{(1.1)} & 51.2 \tiny{(1.0)} \\
&$-hard$                       &  76.5 \tiny{(1.3)}  & 62.2 \tiny{(0.7)} & 70.6 \tiny{(0.5)} & 49.5 \tiny{(0.8)} \\
\bottomrule
\end{tabular} 
}
\caption{Ablation study on Cora and Citeseer datasets.}
\vspace{-0.05in}
\label{ablation}
\end{table}

\noindent\textbf{\emph{RQ3}: Ricci Curvature \& Clustering.}
We discuss why Ricci curvature works.
Empirically, we further study the clustering capability of Ricci curvature comparing with classic concepts ($gCooL$ with refined modularity, \emph{HostPool} with motif conductance and \emph{RicciCom} with Ricci curvature).
We examine the result clusters from microscopic perspective by cluster density and entropy \citep{DBLP:conf/www/LiJT22}.
The density is $\mathbb E_k[\frac{E_k}{V_k(V_k-1)}]$, where $E_k$ and $V_k$ are the number of edges and nodes in cluster $k$.
The entropy is $-\mathbb E_k[\sum\nolimits_c p_k(c)\log p_k(c)]$, where $p_k(c)$ is the  frequency of class (label) $c$ occurred in cluster $k$. Lower entropy means better result, i.e., the cluster contains a major class.
The results are visualized in Fig. 2.
After a few hundred epochs, Ricci methods achieves even better density than moduarity/conductance methods.
\emph{It shows the clustering capability of Ricci curvature, verifying our motivation.}
Also, we have lower entropy than RicciCom.
It is because we further introduce the novel curvature space, supporting fine-grained curvature modeling for graph clustering.

\vspace{-0.1in}
\section{Related Work}

\subsection{Deep Graph Clustering}
In the literature, deep graph clustering methods are roughly divided into $3$ categories regarding the learning paradigm.
1) Reconstructive methods provide supervision signal by recovering graph information,  and generate node clusters by applying or incorporating clustering methods \citep{DBLP:conf/cikm/DuvalM22,DBLP:conf/ijcai/MrabahBK22}.
2) Adversarial methods regulate the generative process by a discriminator in a min-max game \citep{DBLP:conf/www/JiaZ0W19,DBLP:conf/ijcai/0002WGWCG20}.
3) Contrastive methods acquire discriminative representation without labels by pulling positive samples together while pushing negative samples apart \citep{DBLP:journals/corr/abs-2204-08504,S3GC}.
Meanwhile, deep methods are introduced to bipartite graphs \citep{DBLP:conf/cikm/ZhouWZJH22}, signed graphs \citep{DBLP:conf/sdm/HeRWC22,DBLP:conf/icdm/KangLLHK21}, temporal graphs \citep{DBLP:conf/aaai/YaoJ21}, heterogeneous graphs \citep{DBLP:journals/corr/abs-2108-03953}, and etc.
Recently, \cite{DBLP:conf/ijcai/HeLJJH21} present a novel generative model with EM algorithm; \cite{DBLP:conf/wsdm/FettalLN22} introduce a strong matrix optimization framework.
Distinguishing from the prior studies, we rethink the problem of graph clustering from the geometric perspective.


\begin{figure} 
\vspace{-0.15in}
\subfigcapskip=-3pt
\subfigure[Cluster Density on Cora]{
\includegraphics[width=0.48\linewidth]{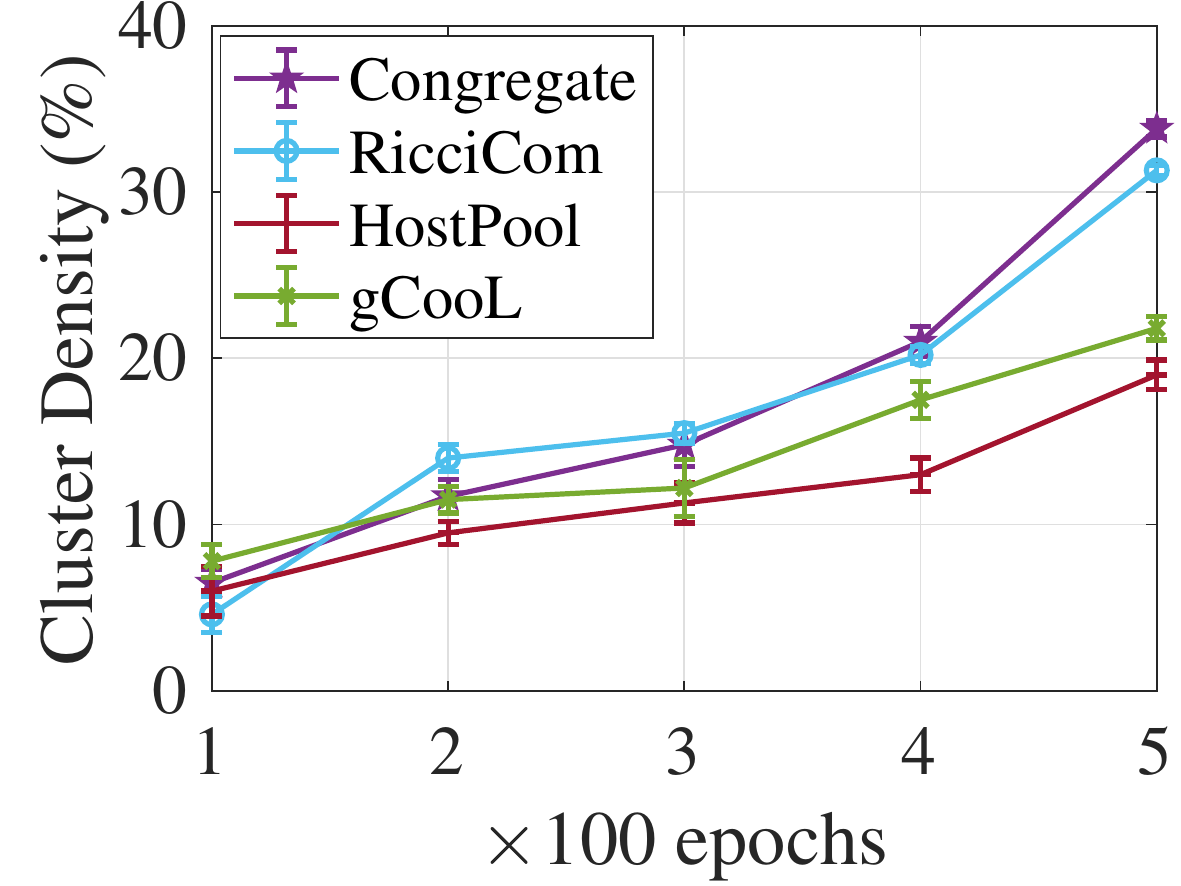}}
\subfigure[Cluster Entropy on Cora]{
\includegraphics[width=0.48\linewidth]{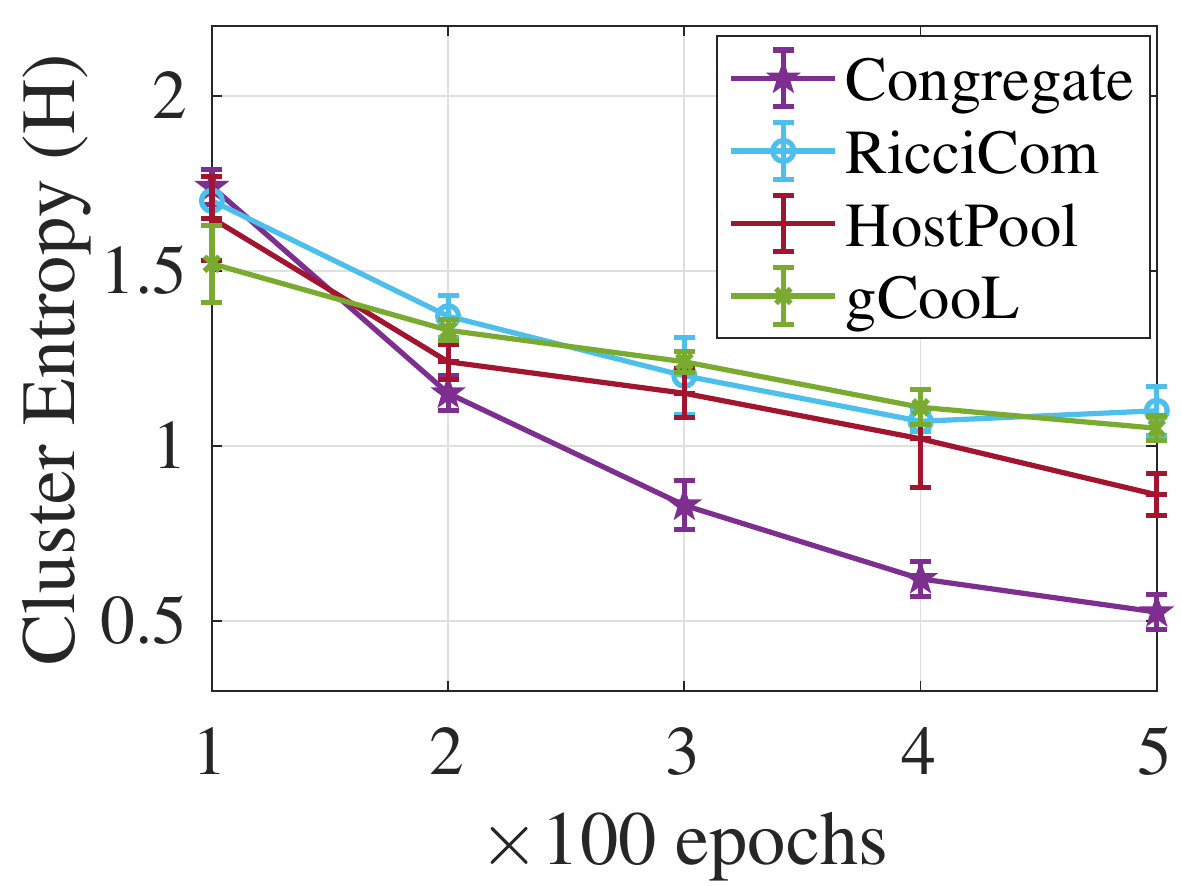}}
\vfill
\subfigure[Cluster Density on Citeseer]{
\includegraphics[width=0.48\linewidth]{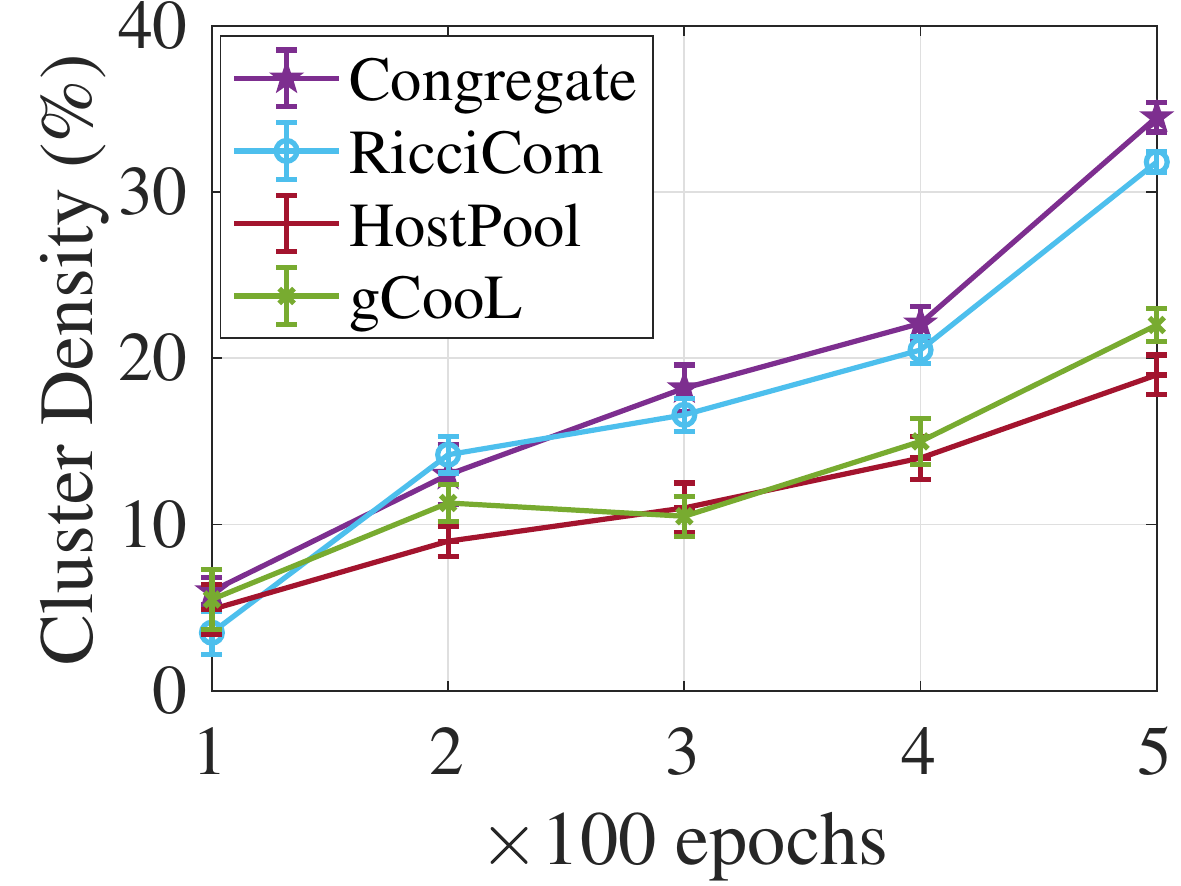}}
\subfigure[Cluster Entropy on Citeseer]{
\includegraphics[width=0.48\linewidth]{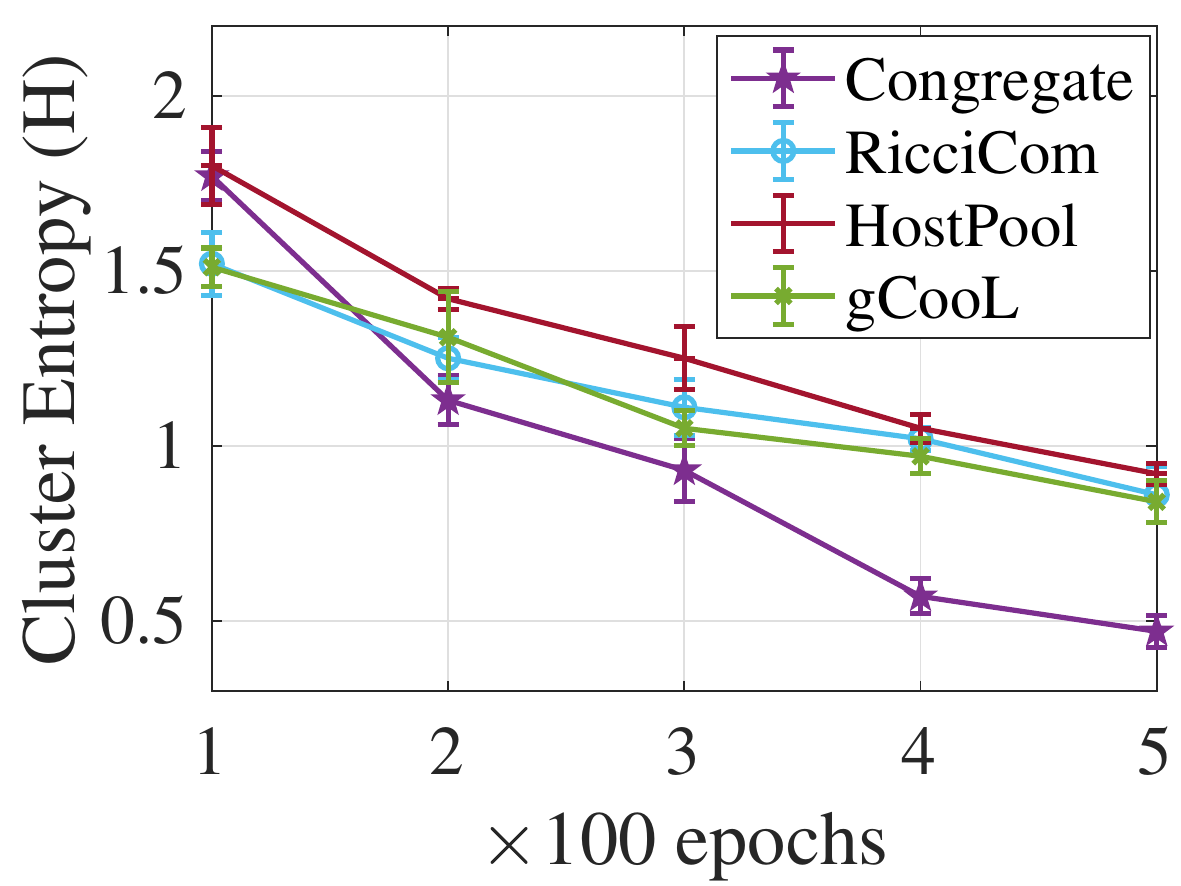}}
\vspace{-0.12in}
\caption{Visualize density and entropy of the clusters.}
\label{case}
\vspace{-0.1in}
\end{figure}

\vspace{-0.05in}
\subsection{Riemannian Graph Learning}

Recent years have witnessed the remarkable success achieved by Riemannian graph learning. 
As hyperbolic space is well aligned with hierarchical or power-law graphs \citep{krioukov2010hyperbolic}, 
shallow models are first introduced \citep{nickel2017poincare,suzuki2019hyperbolic}, and hyperbolic GCNs with different formulations are then proposed \citep{HGCN,HGNN,ZhangWSLS21,HTGN}. 
Beyond hyperbolic space,
$\kappa$-GCN \citep{BachmannBG20} extend GCN to constant-curvature spaces with $\kappa$-sterographical model.
\citet{DBLP:conf/www/YangCPLYX22} model the graph in the dual space of Euclidean and hyperbolic ones.
\citet{DBLP:conf/kdd/XiongZNXP0S22,pRieGNN} study graph learning on a kind of pseudo Riemannian manifold, ultrahyperbolic space.
\citet{NEURIPS2021_b91b1fac} introduce a quotient manifold for graph learning.
\citet{DBLP:conf/aaai/CruceruBG21} study the matrix manifold of Riemannian spaces.
\citet{GuSGR19,DBLP:conf/www/WangWSWNAXYC21} explore node embedding in the product spaces. 
Very recently, \cite{DBLP:journals/corr/abs-2202-01185} investigate in the rotational symmetry of the manifold, but do not consider fine-grained curvature modeling and learnable factors, different from our study.
However, none of existing studies focus on graph clustering in Riemannian manifolds to the best of our knowledge.

\section{Conclusion}

In this paper, we formulate the problem of geometric graph clustering, which is \emph{the first to introduce the curvature space allowing for fine-grained curvature modeling to graph clustering}.
We present an end-to-end \textsc{Congregate} built upon a novel heterogeneous curvature space that we construct for geometric graph clustering with Ricci curvatures.
Accordingly, graph clusters are trained by an augmentation-free contrastive loss, where we pay more attention to both hard positives and hard negatives in our curvature space.
The empirical results show the superiority of our model.

\section{Acknowledgments}
Thanks to the anonymous reviewers.
The authors of this paper were supported in part by National Natural Science Foundation of China under Grant 62202164, the National Key R\&D Program of China through grant 2021YFB1714800,  S\&T Program of Hebei through grant 21340301D and the Fundamental Research Funds for the Central Universities 2022MS018.
Prof. Philip S. Yu is supported in part by NSF under grants III-1763325, III-1909323,  III-2106758, and SaTC-1930941.
Correspond to Li Sun and Hao Peng.





\bibliographystyle{named}
\bibliography{ijcai23}

\begin{thebibliography}{}

\bibitem[\protect\citeauthoryear{Bachmann \bgroup \em et al.\egroup
  }{2020}]{BachmannBG20}
Gregor Bachmann, Gary B{\'{e}}cigneul, and Octavian Ganea.
\newblock Constant curvature graph convolutional networks.
\newblock In {\em Proceedings of ICML}, volume 119, pages 486--496, 2020.

\bibitem[\protect\citeauthoryear{Bo \bgroup \em et al.\egroup
  }{2020}]{DBLP:conf/www/Bo0SZL020}
Deyu Bo, Xiao Wang, Chuan Shi, Meiqi Zhu, Emiao Lu, and Peng Cui.
\newblock Structural deep clustering network.
\newblock In {\em Proceedings of WWW}, pages 1400--1410. {ACM} / {IW3C2}, 2020.

\bibitem[\protect\citeauthoryear{Chami \bgroup \em et al.\egroup }{2019}]{HGCN}
Ines Chami, Zhitao Ying, Christopher R{\'e}, and Jure Leskovec.
\newblock Hyperbolic graph convolutional neural networks.
\newblock In {\em Advances in NeurIPS}, pages 4869--4880, 2019.

\bibitem[\protect\citeauthoryear{Chen and Zhu}{2005}]{2005Ricci}
Bing-Long Chen and Xi-Ping Zhu.
\newblock Ricci flow with surgery on four-manifolds with positive isotropic
  curvature.
\newblock {\em Journal of Differential Geometry}, pages 177--264, 2005.

\bibitem[\protect\citeauthoryear{Chen \bgroup \em et al.\egroup
  }{2022}]{DBLP:conf/acl/ChenHLZLLSZ22}
Weize Chen, Xu~Han, Yankai Lin, Hexu Zhao, Zhiyuan Liu, Peng Li, Maosong Sun,
  and Jie Zhou.
\newblock Fully hyperbolic neural networks.
\newblock In {\em Proceedings of the 60th ACL}, pages 5672--5686. ACL, 2022.

\bibitem[\protect\citeauthoryear{Cruceru \bgroup \em et al.\egroup
  }{2021}]{DBLP:conf/aaai/CruceruBG21}
Calin Cruceru, Gary B{\'{e}}cigneul, and Octavian{-}Eugen Ganea.
\newblock Computationally tractable riemannian manifolds for graph embeddings.
\newblock In {\em Proceedings of AAAI}, pages 7133--7141. {AAAI} Press, 2021.

\bibitem[\protect\citeauthoryear{Cui \bgroup \em et al.\egroup
  }{2020}]{DBLP:conf/kdd/CuiZY020}
Ganqu Cui, Jie Zhou, Cheng Yang, and Zhiyuan Liu.
\newblock Adaptive graph encoder for attributed graph embedding.
\newblock In {\em Proceedings of the 26th {ACM} {SIGKDD}}, pages 976--985.
  {ACM}, 2020.

\bibitem[\protect\citeauthoryear{Dai \bgroup \em et al.\egroup
  }{2021}]{DBLP:conf/cvpr/DaiWGJ21}
Jindou Dai, Yuwei Wu, Zhi Gao, and Yunde Jia.
\newblock A hyperbolic-to-hyperbolic graph convolutional network.
\newblock In {\em Proceedings of CVPR}, pages 154--163, 2021.

\bibitem[\protect\citeauthoryear{Devvrit \bgroup \em et al.\egroup
  }{2022}]{S3GC}
Fnu Devvrit, Aditya Sinha, Inderjit Dhillon, and Prateek Jain.
\newblock {S}$^3${GC}: Scalable self-supervised graph clustering.
\newblock In {\em Advances in 36th NeurIPS}, 2022.

\bibitem[\protect\citeauthoryear{Duval and
  Malliaros}{2022}]{DBLP:conf/cikm/DuvalM22}
Alexandre Duval and Fragkiskos~D. Malliaros.
\newblock Higher-order clustering and pooling for graph neural networks.
\newblock In {\em Proceedings of the 31st CIKM}, pages 426--435. {ACM}, 2022.

\bibitem[\protect\citeauthoryear{Fettal \bgroup \em et al.\egroup
  }{2022}]{DBLP:conf/wsdm/FettalLN22}
Chakib Fettal, Lazhar Labiod, and Mohamed Nadif.
\newblock Efficient graph convolution for joint node representation learning
  and clustering.
\newblock In {\em Proceedings of the 15th {WSDM}}, pages 289--297. {ACM}, 2022.

\bibitem[\protect\citeauthoryear{Giovanni \bgroup \em et al.\egroup
  }{2022}]{DBLP:journals/corr/abs-2202-01185}
Francesco~Di Giovanni, Giulia Luise, and Michael~M. Bronstein.
\newblock Heterogeneous manifolds for curvature-aware graph embedding.
\newblock In {\em Proceedings of the 10th ICLR (GTRL Workshop)}, 2022.

\bibitem[\protect\citeauthoryear{Gong \bgroup \em et al.\egroup
  }{2022}]{DBLP:conf/ijcai/GongZTL22}
Lei Gong, Sihang Zhou, Wenxuan Tu, and Xinwang Liu.
\newblock Attributed graph clustering with dual redundancy reduction.
\newblock In {\em Proceedings of the 31st IJCAI}, pages 3015--3021. ijcai.org,
  2022.

\bibitem[\protect\citeauthoryear{Gu \bgroup \em et al.\egroup }{2019}]{GuSGR19}
Albert Gu, Frederic Sala, Beliz Gunel, and Christopher R{\'{e}}.
\newblock Learning mixed-curvature representations in product spaces.
\newblock In {\em Proceedings of ICLR}, pages 1--21, 2019.

\bibitem[\protect\citeauthoryear{Hassani and Ahmadi}{2020}]{HassaniA20}
Kaveh Hassani and Amir Hosein~Khas Ahmadi.
\newblock Contrastive multi-view representation learning on graphs.
\newblock In {\em Proceedings of ICML}, volume 119, pages 4116--4126, 2020.

\bibitem[\protect\citeauthoryear{He \bgroup \em et al.\egroup
  }{2021}]{DBLP:conf/ijcai/HeLJJH21}
Dongxiao He, Shuai Li, Di~Jin, Pengfei Jiao, and Yuxiao Huang.
\newblock Self-guided community detection on networks with missing edges.
\newblock In {\em Proceedings of the 30th IJCAI}, pages 3508--3514. ijcai.org,
  2021.

\bibitem[\protect\citeauthoryear{He \bgroup \em et al.\egroup
  }{2022}]{DBLP:conf/sdm/HeRWC22}
Yixuan He, Gesine Reinert, Songchao Wang, and Mihai Cucuringu.
\newblock {SSSNET:} semi-supervised signed network clustering.
\newblock In {\em Proceedings of SDM}, pages 244--252. {SIAM}, 2022.

\bibitem[\protect\citeauthoryear{Hui \bgroup \em et al.\egroup
  }{2020}]{DBLP:conf/aaai/HuiZH20}
Binyuan Hui, Pengfei Zhu, and Qinghua Hu.
\newblock Collaborative graph convolutional networks: Unsupervised learning
  meets semi-supervised learning.
\newblock In {\em Proceedings of the 34th AAAI}, pages 4215--4222. {AAAI}
  Press, 2020.

\bibitem[\protect\citeauthoryear{Jia \bgroup \em et al.\egroup
  }{2019}]{DBLP:conf/www/JiaZ0W19}
Yuting Jia, Qinqin Zhang, Weinan Zhang, and Xinbing Wang.
\newblock Communitygan: Community detection with generative adversarial nets.
\newblock In {\em Proceedings of the WWW}, pages 784--794. {ACM}, 2019.

\bibitem[\protect\citeauthoryear{Jost and
  Liu}{2014}]{DBLP:journals/dcg/JostL14}
J{\"{u}}rgen Jost and Shiping Liu.
\newblock Ollivier's ricci curvature, local clustering and curvature-dimension
  inequalities on graphs.
\newblock {\em Discrete \& Computational Geometry}, 51(2):300--322, 2014.

\bibitem[\protect\citeauthoryear{Kang \bgroup \em et al.\egroup
  }{2021}]{DBLP:conf/icdm/KangLLHK21}
Yoonsuk Kang, Woncheol Lee, Yeon{-}Chang Lee, Kyungsik Han, and Sang{-}Wook
  Kim.
\newblock Adversarial learning of balanced triangles for accurate community
  detection on signed networks.
\newblock In {\em Proceedings of ICDM}, pages 1150--1155. {IEEE}, 2021.

\bibitem[\protect\citeauthoryear{Khan and
  Kleinsteuber}{2022}]{DBLP:journals/corr/abs-2108-03953}
Rayyan~Ahmad Khan and Martin Kleinsteuber.
\newblock A framework for joint unsupervised learning of cluster-aware
  embedding for heterogeneous networks.
\newblock In {\em Proceedings of the 15th {WSDM}}. {ACM}, 2022.

\bibitem[\protect\citeauthoryear{Kipf and Welling}{2016}]{kipf2016variational}
Thomas~N Kipf and Max Welling.
\newblock Variational graph auto-encoders.
\newblock {\em NeurIPS Bayesian Deep Learning Workshop}, 2016.

\bibitem[\protect\citeauthoryear{Kochurov \bgroup \em et al.\egroup
  }{2020}]{DBLP:journals/corr/abs-2005-02819}
Max Kochurov, Rasul Karimov, and Serge Kozlukov.
\newblock Geoopt: Riemannian optimization in pytorch.
\newblock In {\em Proceedings of ICML (GRLB Workshop)}. {PMLR}, 2020.

\bibitem[\protect\citeauthoryear{Krioukov \bgroup \em et al.\egroup
  }{2010}]{krioukov2010hyperbolic}
Dmitri Krioukov, Fragkiskos Papadopoulos, Maksim Kitsak, Amin Vahdat, and
  Mari{\'a}n Bogun{\'a}.
\newblock Hyperbolic geometry of complex networks.
\newblock {\em Physical Review E}, 82(3):036106, 2010.

\bibitem[\protect\citeauthoryear{Law}{2021}]{NEURIPS2021_b91b1fac}
Marc Law.
\newblock Ultrahyperbolic neural networks.
\newblock In {\em Advances in Neural Information Processing Systems},
  volume~34, pages 22058--22069, 2021.

\bibitem[\protect\citeauthoryear{Lee}{2013}]{2013manifold}
John~M. Lee.
\newblock {\em Introduction to Smooth Manifolds (2nd Edition)}.
\newblock Springer, 2013.

\bibitem[\protect\citeauthoryear{Li \bgroup \em et al.\egroup
  }{2020}]{DBLP:conf/nips/LiYLZZR0H20}
Jia Li, Jianwei Yu, Jiajin Li, Honglei Zhang, Kangfei Zhao, Yu~Rong, Hong
  Cheng, and Junzhou Huang.
\newblock Dirichlet graph variational autoencoder.
\newblock In {\em Advances in NeurIPS}, 2020.

\bibitem[\protect\citeauthoryear{Li \bgroup \em et al.\egroup
  }{2022}]{DBLP:conf/www/LiJT22}
Bolian Li, Baoyu Jing, and Hanghang Tong.
\newblock Graph communal contrastive learning.
\newblock In {\em Proceedings of The {ACM} Web Conference}, pages 1203--1213.
  {ACM}, 2022.

\bibitem[\protect\citeauthoryear{Lin \bgroup \em et al.\egroup
  }{2011}]{2011Ricci}
Yong Lin, Linyuan Lu, and Shing{-}Tung Yau.
\newblock Ricci curvature of graphs.
\newblock {\em Tohoku Mathematical Journal}, 63(4):605--627, 2011.

\bibitem[\protect\citeauthoryear{Liu \bgroup \em et al.\egroup }{2019}]{HGNN}
Qi~Liu, Maximilian Nickel, and Douwe Kiela.
\newblock Hyperbolic graph neural networks.
\newblock In {\em Advances in NeurIPS}, pages 8228--8239, 2019.

\bibitem[\protect\citeauthoryear{Liu \bgroup \em et al.\egroup
  }{2023}]{DBLP:journals/corr/abs-2212-08665}
Yue Liu, Xihong Yang, Sihang Zhou, Xinwang Liu, Zhen Wang, Ke~Liang, Wenxuan
  Tu, Liang Li, Jingcan Duan, and Cancan Chen.
\newblock Hard sample aware network for contrastive deep graph clustering.
\newblock In {\em Proceedings of the AAAI}, 2023.

\bibitem[\protect\citeauthoryear{Mrabah \bgroup \em et al.\egroup
  }{2022}]{DBLP:conf/ijcai/MrabahBK22}
Nairouz Mrabah, Mohamed Bouguessa, and Riadh Ksantini.
\newblock Escaping feature twist: {A} variational graph auto-encoder for node
  clustering.
\newblock In {\em Proceedings of the 31st IJCAI}, pages 3351--3357. ijcai.org,
  2022.

\bibitem[\protect\citeauthoryear{Ni \bgroup \em et al.\egroup
  }{2019}]{CDRiccFlow}
Chien{-}Chun Ni, Yu{-}Yao Lin, Feng Luo, and Jie Gao.
\newblock Community detection on networks with ricci flow.
\newblock {\em Nature Scientific Reports}, 9(9984), 2019.

\bibitem[\protect\citeauthoryear{Nickel and Kiela}{2017}]{nickel2017poincare}
Maximillian Nickel and Douwe Kiela.
\newblock Poincar{\'e} embeddings for learning hierarchical representations.
\newblock In {\em Advances in NeurIPS}, pages 6338--6347, 2017.

\bibitem[\protect\citeauthoryear{Pan \bgroup \em et al.\egroup
  }{2020}]{DBLP:journals/tcyb/PanHFLJZ20}
Shirui Pan, Ruiqi Hu, Sai{-}Fu Fung, Guodong Long, Jing Jiang, and Chengqi
  Zhang.
\newblock Learning graph embedding with adversarial training methods.
\newblock {\em {IEEE} Trans. on Cybern.}, 50(6):2475--2487, 2020.

\bibitem[\protect\citeauthoryear{Park \bgroup \em et al.\egroup
  }{2022}]{DBLP:journals/corr/abs-2204-08504}
Namyong Park, Ryan~A. Rossi, Eunyee Koh, Iftikhar~Ahamath Burhanuddin, Sungchul
  Kim, Fan Du, Nesreen~K. Ahmed, and Christos Faloutsos.
\newblock {CGC:} contrastive graph clustering for community detection and
  tracking.
\newblock In {\em Proceedings of the Web Conference}, pages 1115--1126, 2022.

\bibitem[\protect\citeauthoryear{Peng \bgroup \em et al.\egroup
  }{2021}]{DBLP:conf/mm/PengLJH21}
Zhihao Peng, Hui Liu, Yuheng Jia, and Junhui Hou.
\newblock Attention-driven graph clustering network.
\newblock In {\em Proceedings of ACM MM}, pages 935--943. {ACM}, 2021.

\bibitem[\protect\citeauthoryear{Robinson \bgroup \em et al.\egroup
  }{2021}]{DBLP:conf/iclr/RobinsonCSJ21}
Joshua~David Robinson, Ching{-}Yao Chuang, Suvrit Sra, and Stefanie Jegelka.
\newblock Contrastive learning with hard negative samples.
\newblock In {\em Proceedings of the 9th ICLR}. OpenReview.net, 2021.

\bibitem[\protect\citeauthoryear{Sia \bgroup \em et al.\egroup
  }{2019}]{ORCurvCD}
Jayson Sia, Edmond Jonckheere, and Paul Bogdan.
\newblock Ollivier-ricci curvature-based method to community detection in
  complex networks.
\newblock {\em Nature Scientific Reports}, 9(9800), 2019.

\bibitem[\protect\citeauthoryear{Skopek \bgroup \em et al.\egroup
  }{2020}]{SkopekGB20}
Ondrej Skopek, Octavian{-}Eugen Ganea, and Gary B{\'{e}}cigneul.
\newblock Mixed-curvature variational autoencoders.
\newblock In {\em Proceedings of ICLR}, pages 1--54, 2020.

\bibitem[\protect\citeauthoryear{Sun \bgroup \em et al.\egroup
  }{2022}]{DBLP:conf/cikm/0008YPY22}
Li~Sun, Junda Ye, Hao Peng, and Philip~S. Yu.
\newblock A self-supervised riemannian {GNN} with time varying curvature for
  temporal graph learning.
\newblock In {\em Proceedings of the 31st CIKM}, pages 1827--1836. {ACM}, 2022.

\bibitem[\protect\citeauthoryear{Suzuki \bgroup \em et al.\egroup
  }{2019}]{suzuki2019hyperbolic}
Ryota Suzuki, Ryusuke Takahama, and Shun Onoda.
\newblock Hyperbolic disk embeddings for directed acyclic graphs.
\newblock In {\em Proceedings of ICML}, pages 6066--6075, 2019.

\bibitem[\protect\citeauthoryear{Veli{\v{c}}kovi{\'{c}} \bgroup \em et
  al.\egroup }{2018}]{velickovic2018graph}
Petar Veli{\v{c}}kovi{\'{c}}, Guillem Cucurull, Arantxa Casanova, Adriana
  Romero, Pietro Li{\`{o}}, and Yoshua Bengio.
\newblock Graph attention networks.
\newblock In {\em Proceedings of ICLR}, pages 1--12, 2018.

\bibitem[\protect\citeauthoryear{Veli{\v{c}}kovi{\'{c}} \bgroup \em et
  al.\egroup }{2019}]{VelickovicFHLBH19}
Petar Veli{\v{c}}kovi{\'{c}}, William Fedus, William~L. Hamilton, Pietro
  Li{\`{o}}, Yoshua Bengio, and R.~Devon Hjelm.
\newblock Deep graph infomax.
\newblock In {\em Proceedings of ICLR}, pages 1--24, 2019.

\bibitem[\protect\citeauthoryear{Wang \bgroup \em et al.\egroup
  }{2019}]{DBLP:conf/ijcai/WangPHLJZ19}
Chun Wang, Shirui Pan, Ruiqi Hu, Guodong Long, Jing Jiang, and Chengqi Zhang.
\newblock Attributed graph clustering: {A} deep attentional embedding approach.
\newblock In {\em Proceedings of the 28th IJCAI}, pages 3670--3676. ijcai.org,
  2019.

\bibitem[\protect\citeauthoryear{Wang \bgroup \em et al.\egroup
  }{2021}]{DBLP:conf/www/WangWSWNAXYC21}
Shen Wang, Xiaokai Wei, C{\'{\i}}cero~Nogueira dos Santos, Zhiguo Wang, Ramesh
  Nallapati, Andrew~O. Arnold, Bing Xiang, Philip~S. Yu, and Isabel~F. Cruz.
\newblock Mixed-curvature multi-relational graph neural network for knowledge
  graph completion.
\newblock In {\em Proceedings of The {ACM} Web Conference}, pages 1761--1771.
  {ACM} / {IW3C2}, 2021.

\bibitem[\protect\citeauthoryear{Xia \bgroup \em et al.\egroup
  }{2022}]{DBLP:conf/icml/XiaWWCL22}
Jun Xia, Lirong Wu, Ge~Wang, Jintao Chen, and Stan~Z. Li.
\newblock Progcl: Rethinking hard negative mining in graph contrastive
  learning.
\newblock In {\em Proceedings of ICML}, volume 162, pages 24332--24346. {PMLR},
  2022.

\bibitem[\protect\citeauthoryear{Xiong \bgroup \em et al.\egroup
  }{2022a}]{DBLP:conf/kdd/XiongZNXP0S22}
Bo~Xiong, Shichao Zhu, Mojtaba Nayyeri, Chengjin Xu, Shirui Pan, Chuan Zhou,
  and Steffen Staab.
\newblock Ultrahyperbolic knowledge graph embeddings.
\newblock In {\em Proceedings of KDD}, pages 2130--2139. {ACM}, 2022.

\bibitem[\protect\citeauthoryear{Xiong \bgroup \em et al.\egroup
  }{2022b}]{pRieGNN}
Bo~Xiong, Shichao Zhu, Nico Potyka, Shirui Pan, Chuan Zhu, and Steffen Staab.
\newblock Pseudo-riemannian graph convolutional networks.
\newblock In {\em Advances in 36th NeurIPS}, pages 1--21, 2022.

\bibitem[\protect\citeauthoryear{Yang \bgroup \em et al.\egroup
  }{2020}]{DBLP:conf/ijcai/0002WGWCG20}
Liang Yang, Yuexue Wang, Junhua Gu, Chuan Wang, Xiaochun Cao, and Yuanfang Guo.
\newblock {JANE:} jointly adversarial network embedding.
\newblock In {\em Proceedings of the 29th IJCAI}, pages 1381--1387. ijcai.org,
  2020.

\bibitem[\protect\citeauthoryear{Yang \bgroup \em et al.\egroup }{2021}]{HTGN}
Menglin Yang, Min Zhou, Marcus Kalander, Zengfeng Huang, and Irwin King.
\newblock Discrete-time temporal network embedding via implicit hierarchical
  learning in hyperbolic space.
\newblock In {\em Proceedings of KDD}, pages 1975--1985. {ACM}, 2021.

\bibitem[\protect\citeauthoryear{Yang \bgroup \em et al.\egroup
  }{2022}]{DBLP:conf/www/YangCPLYX22}
Haoran Yang, Hongxu Chen, Shirui Pan, Lin Li, Philip~S. Yu, and Guandong Xu.
\newblock Dual space graph contrastive learning.
\newblock In {\em Proceedings of The {ACM} Web Conference}, pages 1238--1247,
  2022.

\bibitem[\protect\citeauthoryear{Yao and
  Joe{-}Wong}{2021}]{DBLP:conf/aaai/YaoJ21}
Yuhang Yao and Carlee Joe{-}Wong.
\newblock Interpretable clustering on dynamic graphs with recurrent graph
  neural networks.
\newblock In {\em Proceedings of the 35th AAAI}, pages 4608--4616. {AAAI}
  Press, 2021.

\bibitem[\protect\citeauthoryear{Ye \bgroup \em et al.\egroup
  }{2020}]{DBLP:conf/iclr/YeLM0020}
Ze~Ye, Kin~Sum Liu, Tengfei Ma, Jie Gao, and Chao Chen.
\newblock Curvature graph network.
\newblock In {\em Proceedings of the 8th ICLR}, 2020.

\bibitem[\protect\citeauthoryear{Yin \bgroup \em et al.\egroup
  }{2017}]{DBLP:conf/kdd/YinBLG17}
Hao Yin, Austin~R. Benson, Jure Leskovec, and David~F. Gleich.
\newblock Local higher-order graph clustering.
\newblock In {\em Proceedings of the 23rd {ACM} {SIGKDD}}, pages 555--564.
  {ACM}, 2017.

\bibitem[\protect\citeauthoryear{Zhang \bgroup \em et al.\egroup
  }{2021}]{ZhangWSLS21}
Yiding Zhang, Xiao Wang, Chuan Shi, Nian Liu, and Guojie Song.
\newblock Lorentzian graph convolutional networks.
\newblock In {\em Proceedings of WWW}, pages 1249--1261, 2021.

\bibitem[\protect\citeauthoryear{Zhao \bgroup \em et al.\egroup
  }{2021}]{DBLP:conf/ijcai/ZhaoYWYD21}
Han Zhao, Xu~Yang, Zhenru Wang, Erkun Yang, and Cheng Deng.
\newblock Graph debiased contrastive learning with joint representation
  clustering.
\newblock In {\em Proceedings of the 30th IJCAI}, pages 3434--3440. ijcai.org,
  2021.

\bibitem[\protect\citeauthoryear{Zhou \bgroup \em et al.\egroup
  }{2022}]{DBLP:conf/cikm/ZhouWZJH22}
Cangqi Zhou, Yuxiang Wang, Jing Zhang, Jiqiong Jiang, and Dianming Hu.
\newblock End-to-end modularity-based community co-partition in bipartite
  networks.
\newblock In {\em Proceedings of the 31st CIKM}, pages 2711--2720. {ACM}, 2022.

\end{thebibliography}

\end{document}